\documentclass[11pt]{article}

\usepackage[final]{acl}
\usepackage{siunitx}

\usepackage{times}
\usepackage{latexsym}

\usepackage[T1]{fontenc}
\usepackage[utf8]{inputenc}

\usepackage{microtype}

\usepackage{inconsolata}

\usepackage{graphicx}

\usepackage{tcolorbox}
\usepackage{booktabs}
\usepackage{multirow}
\usepackage{changepage}
\usepackage{amssymb}
\usepackage{amsmath}
\usepackage{pifont}
\usepackage{makecell}
\usepackage[table]{xcolor}
\usepackage{cleveref}
\usepackage{hyperref}

\title{Grid Spatial Understanding: A Dataset for Textual Spatial Reasoning over Grids, Embodied Settings, and Coordinate Structures}
\author{Risham Sidhu, Julia Hockenmaier \\
         University of Illinois at Urbana-Champaign \\ 
}

\begin{document}
\newcommand{\todo}[1]{\textcolor{red}{#1}}

\maketitle
\begin{abstract}
We introduce GSU, a text-only grid dataset to evaluate the spatial reasoning capabilities of LLMs over 3 core tasks: navigation, object localization, and structure composition. By forgoing visual inputs, isolating spatial reasoning from perception, we show that while most models grasp basic grid concepts, they struggle with frames of reference relative to an embodied agent and identifying 3D shapes from coordinate lists. We also find that exposure to a visual modality does not provide a generalizable understanding of 3D space that VLMs are able to utilize for these tasks. Finally, we show that while the very latest frontier models can solve the provided tasks (though harder variants may still stump them), fully fine-tuning a small LM or LORA fine-tuning a small LLM show potential to match frontier model performance, suggesting an avenue for specialized embodied agents.
\end{abstract}

\section{Introduction}
With the advent of more powerful LLMs, their applications have broaden from text generation to complex problems ranging from advanced mathematical reasoning to multimodal agents. Many of these tasks require the ability to reason about space, whether to solve geometry problems \citep{zhao2025geometryproblemsolvinglarge,  gao2025gllavasolvinggeometricproblem} or to understand verbal commands about their surroundings as an embodied agent \citep{zhang2024visionandlanguagenavigationtodaytomorrow, feng2025embodiedaillmsworld}, to complete tasks while interacting with objects in complex environments \citep{padmakumar2021teachtaskdrivenembodiedagents, shridhar2020alfredbenchmarkinterpretinggrounded}, to describe visual scenes \citep{fu2024scenellmextendinglanguagemodel}, or to reason over 2- and 3-D images \citep{cheng2024spatialrgptgroundedspatialreasoning, chen2024spatialvlmendowingvisionlanguagemodels}.

Approaches to such tasks incorporate additional modalities through vision encoders, projections of point cloud data, or more specialized representations, but using these effectively necessitates updates to base LLM architecture or additional training. Furthermore, even with such additions, it is unclear if fundamentally non-physical models that have been trained primarily on unpaired text can truly understand 3D space. Despite recent advances on various spatial reasoning tasks, the internal reasoning processes of LLMs, VLMs, and variations thereof are often not interpretable, meaning that merely measuring performance is distinct from evaluating understanding.
And even when performance is strong, further investigations into how additional modalities, like vision, are used in spatial tasks reveal that multi-modal LLMs receive little additional information from images when provided with rich text representations \citep{wang2024pictureworththousandwords}.

In that case, textual descriptions of physical space may serve as a convenient proxy for complex visual inputs when evaluating spatial reasoning. By removing visual perceptions, they offer a controlled method of probing whether LLMs posses robust internal representations of space, and if such representations can be manipulated with textual descriptions alone. Given that grids are an established representation for 3D space, they are a natural choice to express 3D layouts and movements. 

In this work, we present GSU, a dataset composed of 3 simple text-only grid spatial reasoning tasks aimed at applications in embodied domains, to serve as a diagnostic for the inherent spatial reasoning abilities of existing models and identify where further work is needed to induce the spacial capabilities embodied agents require.

\begin{itemize}
    \item The \textbf{Navigation} Task requires instruction generation or following on 2D and 3D grids with two potential spatial frames of reference: \textit{Cardinal} (i.e. fixed), testing recognition of 1D directional differences in coordiantes, and \textit{Egocentric} (i.e. moving with an embodied agent), testing understanding of rotating/updating spatial reference frames.
    
    \item The \textbf{Object Localization} task asks where targets are relative to a viewer (\textit{Egocentric}), testing 2D and 3D direction differences between coordinates, or relative to a separate spatial reference (\textit{Allocentric}), testing spatial reference in collaborative settings.
    
    \item The \textbf{Structure Composition} task needs models to translate sets of coordinates into basic shapes (e.g. rows, columns, cubes), testing ability to link coordinates to real-world shapes and process large numbers of coordinates.
\end{itemize}

% \todo{	introducing this task for textual / mathematical spatial reasoning 
% 	novelty: 
% 		textual grid coordinates
% 		egocentric frames of reference (relative to an embodied viewer) and relative to reference objects,
% 		comparison of generating vs following instructions, 
% 		composition of basic shapes
% 	utility: 
% 		generalizing parametric spatial information from other modalities, 
% 		representing internal models of space, 
% 		communicating about spatial concepts/information, 
% 		measure of this capability frontier  
% }

\begin{figure*}[h] 
    \centering 
    \includegraphics[width=\textwidth]{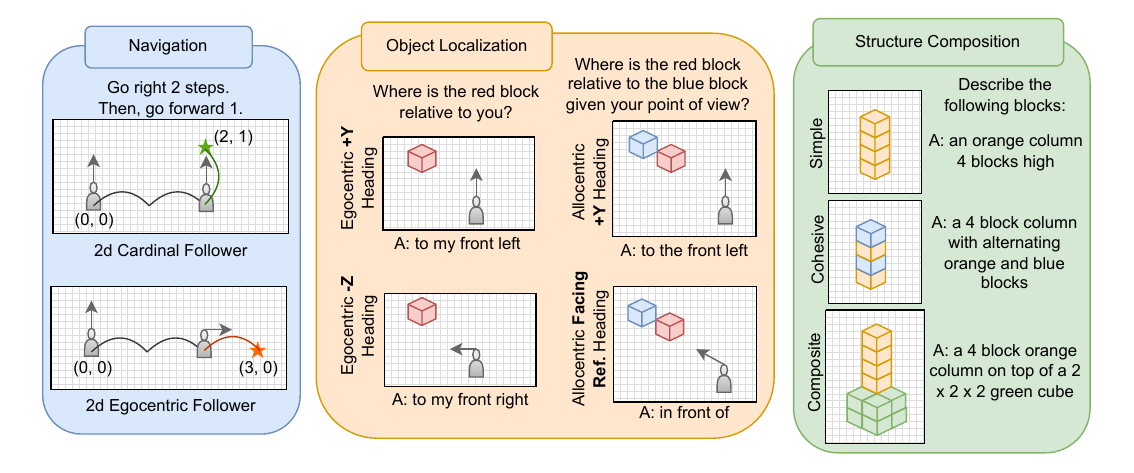}
    \caption{The GSU Dataset tasks and settings visualized for clarity (note that the models do NOT receive these visuals and instead receive a textual version of the environment shown in Appendix \ref{sec:appendix}). The grey arrows indicate the heading, i.e. direction that the embodied agent is facing. In the Navigation task, they remain aligned with +Y in the \textit{Cardinal} setting and rotate to reflect the last direction of travel in the \textit{Egocentric} setting, which affects the direction of the second step and the final coordinates. For the Object Localization task, we show how different headings affect the spatial relations between the target and the reference. For the Structure Composition task, we show the 3 structure categories that models may be asked to describe.} 
    \label{fig:gsu_viz}
\end{figure*}

\section{The GSU Dataset}
We now expand on each of these relatively simple spatial tasks and the specific facets of textual coordinate reasoning they test. For more details on generation and dataset statistics, please refer to Appendix \ref{sec:gen_deets}.

\subsection{The Navigation Task}
The first task measures models' understanding of textual coordinates as a representation of physical space. In this task, models must correlate natural language directions (i.e. "left", "down", "backward") with grid traversals across coordinates differing in 1 dimension at a time, i.e. equating moving from (0, 0) to (2, 0), to (2, 1) with moving "right 2" and "forward 1".

\subsubsection{Instructor vs. Follower}
VLN tasks are generally a conversation between two counterparts: an instructor and a follower. While the instructor is often a person and the follower an agent, this is not necessarily always the case (e.g. a collaborative system with multiple agents, \citet{narayan-chen-etal-2019-collaborative}'s MDC Architect task, voice assistants, etc). Models should thus be proficient in both directions of this mapping between verbal spatial directions and coordinate traversals. 

In the \textit{Follower} setting, models receive a set of natural language instructions explaining how to traverse the grid (e.g. \textit{"move right 2 steps and then forward 1"}) and must output a coordinate reflecting their final location (e.g. $(2, 1)$). Whereas in the \textit{Instructor} setting, models receive a grid traversal expressed as a series of intermediate coordinates (e.g. $(0, 0), (2, 0), (2, 1)$) and must output the natural language instructions that correlate to that path (e.g. \textit{"move right 2 and forward 1"}).

\subsubsection{Spatial Frames of Reference}

We consider two frames of reference for the spatial directions of the grid. In the \textit{Cardinal} setting, the spatial directions are fixed to the axes of the grid and never change, i.e. to move forward always means to increase the y value, while to move left always means to decrease the x value. While many grid based tasks use this fixed perspective of spatial relations (e.g. \citet{aghzal2025largelanguagemodelsgood}), it does not transfer well to practical collaborative settings where an embodied agent or a human it is instructing have their own frame of reference that may not align with a fixed grid. 

The \textit{Egocentric} formulation aligns spatial directions with the perspective of an embodied subject navigating the grid, orienting the spatial directions with their body, and so rotating where they are relative to the fixed axes of the grid. For example, in the \textit{Egocentric} setting to go right 2 steps and then forward 1 step would first entail turning to the right (90$^{\circ}$ clockwise to face $+X$) and moving two steps in that direction. Then, moving 1 step forward would no longer mean increasing the $y$ value, but rather continuing forward in our current heading/direction of travel, i.e. increasing our $x$ value. This would be equivalent to moving 2 steps right and then 1 more step right in the \textit{Cardinal} setting. (See Figure \ref{fig:gsu_viz} for a visualization.) 
% In the 3d settings, however, we do not rotate to move up or down--this is limited to movement in the horizontal plane ($XY$)

Additionally, we include a simpler form of the \textit{Egocentric Instructor} setting, \textit{Card2Ego}. Rather than going from coordinates to egocentric spatial directions, \textit{Card2Ego} asks models to translate cardinal spatial directions (North/South/East/West) into egocentric spatial directions (left/right/forward/backward).
% Noticing how much models struggled with the \textit{Egocentric Instructor} settings, we also formulated one additional setting, \textit{Card2Ego}, which simplifies the \textit{Egocentric Instructor} task from the original (coordinates to spatial directions) into cardinal spatial directions to egocentric spatial directions. For this particular \textit{Instructor}-only formulation, we differentiate between the cardinal and egocentric spatial references by using North/South/East/West for the cardinal inputs and expecting Left/Right/Forward/Back for the egocentric conversion. 

\subsection{The Object Localization Task}

The object localization task focuses on understanding how grid coordinates are situated spatially relative to one another. Specifically, given a viewer at a location and with a particular heading, we ask models to determine where a target at a randomly generated coordinate is relative to that viewer (\textit{Egocentric}) or where it is relative to a reference at another randomly generated coordinate (\textit{Allocentric}). In the former, models are expected to reply with something to the effect of the object is "to \textit{my} right" and in the latter with the target is "to \textit{the} right of" the reference.

In this task, we vary the heading of the viewer (i.e. which axis they are facing parallel to), the distance between the target and the reference (are they adjacent to each other or further away), and add distractors (see Appendix \ref{sec:gen_deets} for more).

\subsection{The Structure Composition Task}
The final task explores a model's ability to understand how blocks/voxels can be composed into larger overall structures. Not only does it require an understanding of how coordinates relate to real space, but also a knowledge of basic shapes/forms (e.g. a series of blocks that differ only in their z-dimension are stacked vertically and thus a column) and the ability to compose large lists of coordinates into meaningful information. 

\subsubsection{Composition Types}
In the most basic setting, \textit{Simple}, the selected coordinates form one structure of a single color and the model must interpret what 3D shape their arrangement forms (e.g. \textit{"an orange row 5 blocks long"} or \textit{"a 3 x 3 x 3 green cube"}). Slightly more complicated is the \textit{Cohesive} setting where multiple colors of blocks form a single structure and the challenge is to understand how those colors interact with the structure (e.g. compare \textit{"a row with alternating orange and blue blocks"} and \textit{"a row with an orange left half and a blue right half"}). Finally, the \textit{Composite} setting places 3 single-colored shapes adjacent to each other, asking the model to identify the 3 shapes, their colors, and where they are relative to each other.

\section{Experiments}
\label{sec:exps}
To explore grid spatial understanding over a variety of model families, sizes, and types, we tested 5 broad categories of models. First, we tested small
text-only models: Mistral-7b (\citet{jiang2023mistral7b}'s Mistral-7b-Instruct-v0.3), Llama-8b (\citet{grattafiori2024llama3herdmodels}'s Llama3.1-8b-Instruct), and Qwen-7b (\citet{qwen2, qwen2.5}'s Qwen2.5-7b-Instruct). 
% We evaluate their performance on the tasks with prompting and ICL, as well as exploring LORA fine-tuning for some tasks. 
Additionally, we prompted their most similarly sized Vision-Language counterparts to see if exposure to a visual modality (even if it is not being utilized) can induce further spatial understanding: Pixtral-12b \citep{agrawal2024pixtral12b}, LlamaVL-11b (\citet{grattafiori2024llama3herdmodels}'s Llama3.2-11b-Vision-Instruct), and QwenVL-7b (\citet{qwen2.5-VL, Qwen-VL, Qwen2VL}'s Qwen2.5-VL-7B-Instruct). 
We also evaluated larger models, as they are known to be more powerful than their smaller counterparts: GPT-OSS20b \citep{openai2025gptoss120bgptoss20bmodel}, Qwen32b and QwenVL-32b (\cite{qwen3technicalreport}'s 32b Instruct variants), Olmo-32b \citep{olmo2025olmo3}, Qwen-72b (\citet{qwen2, qwen2.5}'s Qwen2.5-72B-Instruct). 
Finally, to consider frontier models, we choose GPT-4o \citep{openai2024gpt4technicalreport} and Gemini-3-Pro \citep{geminiteam2025geminifamilyhighlycapable} accessed as Gemini-3.1-Pro-Preview, and to compare overall performance to a much smaller model (<1B params) that can be fully fine-tuned, we fully train FlanT5-large \citep{https://doi.org/10.48550/arxiv.2210.11416}.

For the purposes of prompting we generate test sets for each task of size 100 and for training we generate another of size 3000 (we did not see marked improvement in training by increasing the dataset size beyond this, see Appendix \ref{sec:training_deets} for more details). The \textit{Navigation} and \textit{Object Localization} experiments in Table \ref{tab:all_res} are 1-shot prompting with a handwritten example that includes the full reasoning needed to generate an answer. For the Structure Composition task in Table \ref{tab:all_res} the experiments are 3-shot prompting with synthetic examples and no reasoning. These setting were chosen based on performance (see Tables \ref{tab:fsnr} and \ref{tab:fsnr2} in the Appendix), though we explain the difference in behavior as: the \textit{Navigation} and \textit{Object Localization} tasks require some amount of reasoning to work out headings and spatial relations, thus benefit more from one dense example that gives an example reasoning trace, while the \textit{Structure Composition} task needs no explicit reasoning (just the ability to compose coordinates together) and thus benefits from a larger quantity of examples.

\subsection{Metrics}
\subsubsection{Navigation Metrics}
For the \textit{Follower} task, we evaluate the accuracy of the predicted final coordinates as an exact match (i.e. there is no partial credit for getting the x-coordinate coordinate correct if the y-coordinate in incorrect) and for the \textit{Instructor} we measure the exact match accuracy of the generated instruction chain (i.e. every step must be correct for the model to get credit). For a less exacting metric, we computed the distance between the true final coordinates and: the generated coordinates in \textit{Follower} and the final coordinates obtained by executing the generated instruction chain in \textit{Instructor}. As they tracked closely with the accuracies they are reported in see Appendix Table \ref{tab:dists}. 

\subsubsection{Object Localization Metrics}
\label{sec:spatial_overlap}
To measure model performance on the Object Localization tasks, we compute, $O_s$, the term overlap of the spatial relations in the predicted text, $p$, and the ground truth text, $g$.

$$ O_s = \frac{E_r(p) \cap E_r(g)}{E_r(p) \cup E_r(g)} * 100$$ where the function $E_r$ creates a set from each individual spatial relation extracted from the text. Essentially, it takes how many of the spatial terms in the generated sequence $p$ are correct and divides this by the size of the set union of all correct spatial terms in the ground truth, $g$, and all generated spatial terms in $p$. For example, if the correct spatial relations, $E_r(g)$ are \{"front", "left"\}, the model may score a 0 ($E_r(p)$ = \{"right"\}), a 33.33 ($E_r(p)$ = \{"front", "above"\}), a 50 ($E_r(p)$ = \{"front"\}), a 66.67 ($E_r(p)$ = \{"above", "front", "left"\}), or a 100 ($E_r(p)$ = \{"front", "left"\}). This penalizes models for predicting incorrect terms, while reducing this penalty for harder examples with more spatial relations to account for.

\subsubsection{Structure Composition Metrics}
We continue to use the spatial overlap metric, $O_s$, described in \ref{sec:spatial_overlap} and add three additional overlap metrics: $O_c$, which measures the overlap of the color terms, $O_n$, which measures numerical overlap, and $O_f$, which measures the overlap of the structure formation terms.
For $O_n$, we use the same simple overlap as $O_s$, dividing how many correct numbers were generated by the size of the union of all the ground truth numbers and all generated numbers. However, unlike the Object Localization task, we have the opportunity for multiple structures to have the same color in the \textit{Composite} setting, so the formulation for $O_c$ is slightly different:

$$O_c = \frac{\sum_{\mathit{colors}} \min(C_c(p, \mathit{color}), C_c(g, 
\mathit{color}))}{\sum_{\mathit{colors}} \max(C_c(p, \mathit{color}), C_c(g, \mathit{color}))} * 100$$ where $C_c(\mathit{text}, 
\mathit{term})$ counts the occurrence of $\mathit{term}$ or its synonyms in the text $\mathit{text}$. Thus, a generation that only identifies one yellow shape in a structure with 2 yellow sub-structures is penalized, as is a model that predicts 2 yellow shapes in a structure with just 1. 

$O_f$ follows the same base logic as $O_c$, but with more varied synonyms checked in $C_f$ and partial credit afforded for predicting similar but not correct shapes. E.g. rather than getting no credit for predicting \textit{"a blue row on the ground"} for a blue column, we award partial credit based on how similar the shapes are. (see Appendix \ref{sec:shape_metric}) 
\begin{table*}[h!]
\centering
\small
\begin{tabular}{l|c|c|c|c|c||c|c||c|c|c|c}
\toprule
& \multicolumn{5}{c||}{\textbf{Navigation}} &
\multicolumn{2}{c||}{\textbf{Object Localization}} & \multicolumn{4}{c}{\textbf{Structure Composition}} \\
%\midrule
%\textit{reported as:} & \multicolumn{5}{c||}{\textit{accuracy (out of 1.0)}} & \multicolumn{2}{c||}{\textit{Spatial Overlap} $O_s$} & $O_s$ & $O_c$ & $O_f$ & $O_n$ \\ 
\midrule
& \multicolumn{2}{c|}{\textit{Follower}} & \multicolumn{3}{c||}{\textit{Instructor}} & \textit{Ego-} & \textit{Allo-} & \multirow{2}{*}{\scalebox{1.3}{$O_s$}} & \multirow{2}{*}{\scalebox{1.3}{$O_c$}} & \multirow{2}{*}{\scalebox{1.3}{$O_f$}} & \multirow{2}{*}{\scalebox{1.3}{$O_n$}} \\
& \textit{2D} & \textit{3D} & \textit{Card.} & \textit{Ego.} & \textit{Card2Ego} & \textit{centric} & \textit{centric} & & & \\

%& \multicolumn{2}{c|}{\textbf{Ego. Follower}} & \multicolumn{3}{c||}{\textbf{2D Instructor}} & \multirow{2}{*}{\textbf{Ego.}} & \multirow{2}{*}{\textbf{Allo.}} & \multirow{2}{*}{$O_s$} & \multirow{2}{*}{$O_c$} & \multirow{2}{*}{$O_f$} & \multirow{2}{*}{$O_n$} \\
%& \textbf{2D} & \textbf{3D} & \textbf{Card.} & \textbf{Ego.} & \textbf{Card2Ego} & & & & &\\
%& \rotatebox{75}{Egocentric 2D} & \rotatebox{75}{Egocentric 3D} & \rotatebox{75}{Cardinal 2D} & \rotatebox{75}{Egocentric 2D} &\rotatebox{75}{Card2Ego 2D} & \rotatebox{75}{Egocentric} & \rotatebox{75}{Allocentric}\\
%& \textbf{2D} & \textbf{3D} & \textbf{Card.} & \textbf{Ego.} & \textbf{Card2Ego} & \textbf{Ego.} & \textbf{Allo.} & & & \\
%& acc & acc & acc & acc & acc & $O_s$ & $O_s$& $O_s$ & $O_c$ & $O_f$ & $O_n$ \\ 

\midrule

Mistral-7b & 0.30 & 0.38  & 0.59  & 0.15 & 0.27 & 30.32 & 46.50 &  49.75 & 72.60 & 30.21 & 20.01 \\
\rowcolor{cyan!20} Pixtral & 0.33 & 0.50 & 0.87 & 0.26 & 0.35 & 40.83 & 48.18 & 66.25 & 87.68 & 42.16 & 55.43\\

Llama-8b & 0.33 & 0.47 & 0.69 & 0.06 & 0.09 & 46.17 & 50.18 & 62.04 & 86.69 & 22.70 & 37.75\\
\rowcolor{cyan!20} LlamaVL-11b & 0.32 & 0.47 & 0.74 & 0.14 & 0.28 & 48.40 & 48.08 & 64.48 & 86.58 & 44.44 & 44.05 \\

Qwen-7b & 0.09 & 0.19 & 0.76 & 0.21 & 0.36 & 52.81 & \textbf{54.78} & 54.58 & 85.43 & 28.57 & 42.97 \\
\rowcolor{cyan!20} QwenVL-7b & 0.29 & 0.21 & 0.44 & 0.21 & 0.30 & 48.67 & 44.95 & 54.58 & 85.43 & 33.46 & 43.57\\

GPT-oss-20b & 0.15 & 0.16 & 0.26 & 0.10 & 0.58 & 59.90 & 45.46 & 62.88 & 57.11 & 43.60 & 48.47\\
Olmo-32b & 0.34 & 0.40 & 0.48 & 0.14 & 0.62 & \textbf{85.98} & 54.17 & 65.33 & 91.97 & 48.60 & 54.60\\

Qwen-32b & 0.37 & 0.32 & 0.58 & 0.21 & 0.60 &  33.95 & 27.25 & 69.10 & 87.98 & 42.76 & 66.65\\
\rowcolor{cyan!20} QwenVL-32b & 0.38 & 0.52 & 0.12 & 0.17 & \textbf{0.64} & 68.03 & 50.80 & 67.40 & \textbf{93.62} & \textbf{55.09} & 68.22\\

Llama-70b & 0.41 & 0.65 & 0.26 & 0.09 & 0.1 & 75.23 & 22.14 & 62.17 & 86.88 & 47.80 & 50.95 \\

Qwen-72b & \textbf{0.49} & \textbf{0.75} & \textbf{1.0} & \textbf{0.47} & 0.53 & 79.53 & 53.83 & \textbf{70.67} & 88.11 & 42.81 & \textbf{69.23}  \\
\midrule

GPT-4o & 0.71 & 0.84 & 0.89 & 0.67 & 0.93 & 96.00 & 65.50 & 69.20 & 88.63 & 52.85 & 82.28\\
Gemini-3-Pro & \textbf{1.0} & \textbf{1.0} & \textbf{1.0}  & \textbf{1.0} & \textbf{1.0} & \textbf{100} & \textbf{93.0} & \textbf{79.67} & \textbf{97.83} & \textbf{87.45} & \textbf{90.68} \\ 

\midrule

\rowcolor{orange!5} FlanT5 & 0.80 & 0.82 & 0.99 & \textbf{0.95} & \textbf{1.0} & 19.5 & \textbf{94.5} & 99 & 96.17 & 88.83 & \textbf{97.07} \\
\rowcolor{orange!5} Llama-8b & \textbf{0.82} & \textbf{0.96} & 0.57 & 0.61 & 0.99 & \textbf{87.23} & 81.83 & \textbf{99.33} & \textbf{100} & \textbf{93.42} & 96.33 \\ 
\bottomrule
\end{tabular}

\caption{3 GSU task results across 3 model categories (prompted LLMs and \colorbox{cyan!20}{VLMs}, prompted frontier models, and \colorbox{orange!5}{fine-tuned models}) with best scores from each bolded. \textit{Navigation} results are accuracies: final coordinate accuracies for the \textit{Egocentric Follower} and instruction sequence accuracy for the \textit{2D Instructor}, while \textit{OL} and \textit{SC} results are term overlaps: spatial $O_s$, color $O_c$, shape/form $O_f$, and numerical $O_n$. Other settings in Appendix \ref{sec:sup}. %\textit{Navigation} and \textit{Object Localization} experiments are 1-shot prompting with handwritten reasoning in the example, while \textit{Structure Composition} is Few-Shot prompted with no reasoning. Both setting chosen based on performance, see Appendix \ref{sec:sup} Tables \ref{tab:fsnr} and \ref{tab:fsnr2} for more.}
}

\label{tab:all_res}
\end{table*}

\section{Results and Discussion}
In this section we will briefly cover the 3 GSU task results shown in Table \ref{tab:all_res} before covering broader trends: manipulating spatial reference frames (\ref{sec:msrf}), effects of a visual modality (\ref{sec:vm}), frontier model capability (\ref{sec:frontier}), and improving small LLMs (\ref{sec:llama_lora}).

\label{sec:nav_res}
\textbf{Navigation:} We see that the \textit{Instructor} setting provides a greater challenge for prompted models, with the \textit{2D Egocentric Instructor} in Table \ref{tab:all_res} generally showing significant drops compared to the \textit{2d Egocentric Follower}. The \textit{Egocentric} variant is also more difficult as we see in the accuracy difference between the \textit{Cardinal} and \textit{Egocentric Instructor}. (These patterns hold over all settings, see Appendix \ref{sec:sup} for more details.)

\label{sec:ol_res}
\textbf{Object Localization:}
In the \textit{Egocentric} case we see that smaller models receive scores below 50, consistent with further analysis that they often generate spatial relations for axes in which the target and reference do not differ. Larger models adeptly identify the reasoning structure from the example and isolate which axes coordinates differ on, but then fail to assign the correct spatial relation.

\begin{figure}[h] 
    \centering 
\includegraphics[width=0.9\linewidth]
    %, trim = {0cm 1cm 0cm 2cm}]
    {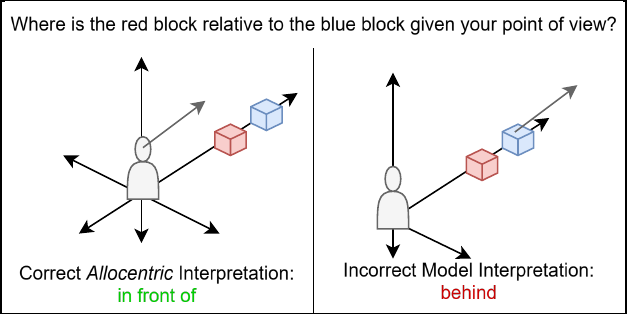}
    \caption{\textit{Allocentric} front/back flipping} 
    \label{fig:qual}
\end{figure}

Most models perform worse on the \textit{Allocentric} variant, suggesting that models struggle to reason about how two grid coordinates are situated relative to each other from the agent perspective (a third coordinate). Those that performed better were likely benefiting from the consistent $+Y$ heading in the \textit{Allocentric} setting rather than displaying a particular strength at the task itself. Notably, many models flip the forward and backward directions as their main source of error in this variant. If the target is between the viewer and the reference, many LLMs describe it as "behind" the reference, rather than "in front" of the reference, as if the reference is the viewer. (see Figure \ref{fig:qual}). 

\label{sec:sc_res}
\textbf{Structure Composition:} In this task, we see similar spatial ($O_s$) performance to the \textit{OL} task, with models often opting to not describe where shapes were relative to each other unless they are certain, and strong color ($O_c$) performance, with mistakes due to omissions rather than hallucinations. The numerical overlap ($O_n$) errors are caused by models miscounting or due to unconventional ways of breaking up structure descriptions: e.g. rather than describing a 3 x 3 x 8 rectangular prism with those dimensions, one generation describes it as a group of 10 columns each 7 blocks high. While this could describe a similar shape (though it could also be a 7 x 10 plane) and certainly accounts for a similar number of blocks, it receives no credit for either metric.

Overall, the generated descriptions are comprehensible and seem reasonable when viewed independently of the ground truth. In 0-shot settings (Appendix Table \ref{tab:old_ol}), models generated vague answers, e.g.: \textit{"a continuous layer of blocks"} rather than concrete shapes, e.g. \textit{"a 4 x 7 rectangle"}. This behavior extends into the few-shot setting, with numerous productions of "structure", "formation", and "platform" modified by rich adjectives, such as \textit{"a large, flat, expansive structure"}. Models also flatten 3D shapes, outputting "square" for cube and "rectangle" for rectangular prisms. Other times they name a shape incorrectly, but still display insight into the formed structure, such as calling a 2 x 2 x 8 rectangular prism a \textit{"cube with height 8 and width 2"} (fundamentally not a cube, but still displaying spatial understanding that is rewarded by the $O_n$ metric instead). 
A notable threshold that stronger models cross is distinguishing rows and columns: stronger models identify which axes the blocks differ on and rarely conflate the two, instead mixing up cubes with rectangular prisms or vertical rectangular prisms with columns. Generally, weaker models also omit reasoning for the shapes they produce, while stronger models like Gemini explain how the coordinate ranges influence its productions, which is reflected in its scores.

\subsection{Manipulating Spatial Reference Frames}
\label{sec:msrf}
The reasoning traces of the models corroborate the score disparity between \textit{Cardinal} and \textit{Egocentric Navigation} tasks in Tables \ref{tab:all_res} and \ref{tab:old_nav}. In the former setting, most models output neat, contained logic with only occasional errors,
%(The \textit{Cardinal Follower} in Appendix Table \ref{tab:old_nav} also performs worse than its \textit{Egocentric} counterparts in Table \ref{tab:all_res}). 
while in the letter, models fail to keep the $X$ and $Y$ axes oriented consistently, leading to generations such as \textit{"we are facing $-Y$ with $+Y$ to our right"} and \textit{"we turned towards $+X$, so we're facing $-X$"}. Larger models, though less likely to make these mistakes, calculate updated headings after a turn and then "forget" to use them in the next step, outputting logic and final coordinates consistent with the fixed grids of the \textit{Cardinal} setting. 

In the \textit{Egocentric Object Localization} tasks, this aversion to non-standard spatial reference frames manifests as non sequitur logic where the model establishes the correct heading in the premise but completes the generation with an incorrect deduction. These deductions are generally consistent with the standard $+Y$ heading, e.g. while facing $-Y$: \textit{"I know that $-X$ is to my right} [correct premise] \textit{"so x values smaller than me are to my left"} [incorrect deduction consistent with a $+Y$ heading instead]. The spatial term overlap, $O_s$ is also higher for the $+Y$ heading than any other (Appendix Table \ref{tab:oo_heading}). This phenomenon might also explain poor performance on the \textit{Allocentric} variant in Table \ref{tab:all_res}, particularly the forward/backing flipping described in Figure \ref{fig:qual}, as models struggle to reconcile how two frames of reference interact with each other.

We suspect that the inability to correctly reorient the spatial reference frame is due in part to exposure to the standard Cartesian orientation in pre-training and in part due to the complexity of updating it. This is supported by the Table \ref{tab:all_res} \textit{Navigation Card2Ego} results, where any mathematical reasoning about textual grid coordinates is completely absent, replaced by converting Cardinal directions into relative ones. LLMs have an easier time with \textit{Card2Ego} than the \textit{Egocentric Instructor} but still fall short, showing that, in addition to struggling with grid coordinates, LLMs consistently struggle with the concept of rotating spatial frames of reference outside of explicit grids too. 

%This difficulty with \textit{Egocentric} reasoning also leads to sharper performance drops as sequence lengths increase, with weaker models reaching almost 0 accuracies by step 4 and even stronger models deteriorating from near perfect performance to numbers near 0.6. This is compounded with the \textit{Instructor} formulation, which is also generally more difficult for models, where we could see earlier steps in sequences with much higher accuracies than later ones and step $i$ in a shorter sequence having higher accuracy than step $i$ in a longer one. \todo{see appendix for more}

\subsection{Effects of the Visual Modality}
\label{sec:vm}
As shown by the 4 VLMs shown in light blue under their LLM counterparts in Table \ref{tab:all_res}, there is no clear correlation between exposure to a visual modality and improved reasoning over textual coordinate grids, especially in the smaller models. For example, on the \textit{Navigation} tasks, Pixtral outperforms Mistral, while Qwen outperforms Qwen-VL, and both Llama models perform similarly, though perhaps the consistent pattern within each model family can be attributed to how vision is integrated into their respective VLMs. In some of the tasks (e.g. \textit{Object Localization}), we see that Qwen-32B-VL outperformed Qwen-32B and in others (i.e. \textit{Structure Composition}), all VLMs outperform their LLM counterparts: these patterns may suggest that VLMs need at least a certain LLM size before being able to utilize information gained from interacting with a textual modality or that they confer more benefits when reasoning over coordinate structures rather than spacial relations.

\subsection{Frontier Capabilities}
\label{sec:frontier}
As expected, the frontier models outperform smaller LLMs, though notably Qwen-72b is able to match their performance on some tasks and even outperform GPT-4o. While proficient in the \textit{Cardinal} setting, GPT-4o struggles to keep track of spatial references in the \textit{Egocentric} one, incorrectly computing the current heading or forgetting to act according to an updated heading, especially in longer sequences (see Appendix Table \ref{tab:models_sequence_length}). For example, GPT-4o generates that it was \textit{"currently facing $-X$, but need[ed] to move right [and moving] right means turning 90 degrees clockwise so that's $+X$"}. This final heading should have been $+Y$, a rotation that GPT handles successfully in shorter sequences. However, a right turn leading to facing $+X$ is consistent with the fixed \textit{Cardinal} frame of reference, which the model seems to default to, even in the midst of egocentric-compliant reasoning.

The latest Gemini model solves multiple tasks completely, even harder variants of the tasks such as arbitrarily long navigation sequences with non-integer step lengths or non-axis aligned turns. It also performs surprisingly well on distinguishing independent block structures from scrambled lists of individual blocks. Given this, we must acknowledge that Gemini-3-Pro has spatial understanding when it comes to spatial frames of reference relative to an embodied agent and composing blocks into larger structures. At the very least, it must have been exposed to similar problems, perhaps in a mathematical context. 

However, that is not to say there is no room for improvement within the domain of textual grid based tasks. We further prompt Gemini on a task combining each of the 3 task categories: an egocentric agent must follow navigation instructions and then answer where two block-based structures are relative to each other (see \ref{sec:combo_task}). This requires the model to: (a) successfully complete egocentric navigation and identify its final heading, (b) distinguish the target and reference structures from distractor objects, all presented as a single list of blocks, and (c) identify where the target is relative to the reference. Here, we find that even Gemini fell short, with a final spatial overlap $O_s$ of 0.62: it sometimes mixes up similar shapes, struggles with the forward/backward issue, and makes slight errors in relative spatial relations given the 3D nature of block structures.

\subsection{Improving Small LLMs}
\label{sec:llama_lora}
We anticipate that more frontier models will develop enough spatial reasoning to solve the base tasks, and eventually the harder variants as well. However, the simplicity of GSU and the subsequent gap between it and real-world scenarios, motivates us to want stronger small LLMs. To that end, we train our worst performing small LLM (Llama-8b) with LORA to elicit an underlying spatial understanding.
Initial experiments reduced performance on the \textit{Navigation} task and so to better match natural LLM outputs, we generate synthetic reasoning traces to accompany the ground truth answers for the \textit{Navigation} and \textit{Object Localization} tasks. (This also matched our prompting findings in Appendix Tables \ref{tab:fsnr} and \ref{tab:fsnr2} that showed a single example with strong reasoning improved performance more than multiple examples without reasoning.) As we show in Table \ref{tab:all_res}, with these reasoning traces, Llama-8b shows improved performance on the \textit{Object Localization} tasks (the reasoning traces offered slight LORA performance boosts). It also more than doubles its performance on the \textit{Navigation Follower} tasks after LORA, outperforming GPT-4o and almost matching Gemini. For the \textit{Navigation Instructor} tasks, improvement is even more drastic for the \textit{Egocentric} and \textit{Card2Ego} versions it initially struggled with, increasing performance tenfold. With this, we show that textual spatial reasoning is not out of reach for smaller models and their potential to match frontier models. %With other training paradigms, we hope that smaller models might even be able to match the strongest frontier models on these tasks.

We also note that fully finetuning a small model like FlanT5 (<1B params) is sufficient to outperform many LLMs and match performance with some frontier models, especially on sequence generation tasks like \textit{Navigation Instructor} (Table \ref{tab:all_res}). This highlights how simple the tasks should be to solve: a lightweight model a fraction of the size is better at identifying shapes from coordinate lists of blocks, where those shapes are relative to each other, and generating instructions for grid navigation.

\section{Related Works}
\textbf{Spatial Reasoning}
%Before the widespread integration of visual modalities, LM spatial reasoning was evaluated through dense text descriptions where all spatial relations were relative to other objects \citep{rizvi-etal-2024-sparc}.
Instead of text, physical information is generally conveyed to LLMs via images, either from external POVs like observing scenes from above \citep{li2024topviewrsvisionlanguagemodelstopview} or from the perspective of embodied agents for applications like VLN, and pointcloud embeddings \citep{hong20233dllminjecting3dworld}. However, even after integrating additional modalities and achieving strong performance on many visual tasks, VLMs could not understand or reason about physical space as capably as humans, struggling with spatial VQA, particularly object orientations in images \citep{liao2024reasoningpathsreferenceobjects, liu2023visualspatialreasoning} and mental transformations like rotations \citep{stogiannidis2025mindgapbenchmarkingspatial}. They fall well short of human performance localizing objects within scenes and reasoning about object dimensions \citep{10654928}. Some have proposed addressing these spatial reasoning shortcomings with training datasets that align 2D images with textual tasks about the orientation of, distance between, and location of nodes \citet{tang2025sparklemasteringbasicspatial}, while others have questioned what VLMs are actually perceiving in images and found that they perform just as well with rich text descriptions instead of input images \citep{wang2024pictureworththousandwords, fu2024blinkmultimodallargelanguage}.

Combined with the difficulty of isolating spatial performance from visual perception, this motivates text-only spatial reasoning to evaluate the "mental" models LLMs have of space. Some use graphs, exploring graph traversals in which LLMs hallucinate nodes, take suboptimal paths, and deteriorate on denser graphs \citep{momennejad2023evaluatingcognitivemapsplanning} or describing spatial relations between graph nodes using multihop reasoning over intermediary nodes \citep{shi2022stepgamenewbenchmarkrobust, li2024advancingspatialreasoninglarge}. Others are nominally set in grids, such as \citet{yamada2024evaluatingspatialunderstandinglarge} which explores 2D grid navigation but only on closed loop paths using relative descriptions of positions and \citet{mayer2025ivisparinteractivevisualspatial}, which proposes a visual puzzle game on grid, but compares visual representation to a chess-style notation for the textual representation. Few consider coordinates for text-only spatial reasoning tasks (a notable exception being \citet{aghzal2025largelanguagemodelsgood}'s evaluation of LLM path planning through obstacles on fixed 2D grids) despite coordinate grids being a standard for representing space and likely plentiful in LLM pretraining data. 

One context in which LLMs may have encountered coordinates or reasoning about them is math problems. While most coordinates in such problems are paired with text and the associated reasoning is unrelated to spatial relations, the success in mathematical reasoning \citep{shao2024deepseekmathpushinglimitsmathematical, ahn2024largelanguagemodelsmathematical, wang2025evaluationllmsmathematicalproblem} and a variety of math datasets \citep{ hendrycks2021measuringmathematicalproblemsolving, chen2022unigeounifyinggeometrylogical} suggests that coordinate spatial reasoning is not outside the realm of LLM abilities.

\textbf{Applications}
GSU distills the spatial reasoning required in embodied settings to its simplest form. As such, the main applications of this work are in embodied AI, from VLN \citep{thomason2019visionanddialognavigation} to cooperative task completion \citep{yang2025embodiedbenchcomprehensivebenchmarkingmultimodal}. While there are several intermediate steps before we can extend our findings to multimodal models, virtual tasks may benefit more directly. Take the MDC \citep{narayan-chen-etal-2019-collaborative}, in which two agents collaborate to build a block-based structure in a 3D grid. Existing approaches struggle with shifting spatial reference frames caused by agents moving around the environment during the course of building (and thus might benefit from methods that improve performance on our \textit{Egocentric Navigation} task). The \textit{Object Localization} task addresses the issue of generating spatial relations for blocks relative to existing structures while the \textit{Structure Composition} aligns with the MDC Architect describing a target structure, a part of the task that remains largely unexplored as recent work has focused on the instruction-following Builder \citep{jayannavar2025bapv2enhancedtask}. %Perhaps because like our \textit{Follower Navigation} it is easier for models than its \textit{Instructor} counterpart. 

Our focus on smaller or unimodal models is further motivated by the practical constraints of real-world embodied AI, where the latencies of frontier models can significantly hinder a physical agent \citep{11394729, pollini2025reducinglatencyllmbasednatural}, hardware constraints may require lightweight models that can be fully housed inside a small robot, privacy concerns may necessitate local models or eschewing multimodal inputs at times \citep{marchang2022assistive, 10.1145/1413862.1413863}, and specialized tasks may need further training anyway.

\section{Conclusion}
In this work, we propose GSU, a novel evaluation framework providing a controlled environment to test spatial reasoning and spatial communication without confounding factors like visual perception. Evaluating a variety of models, we found that most, even some frontier models, struggle to track rotating frames of spatial reference and understand how blocks at grid coordinates may be composed into structures, though recent models are crossing this threshold of spatial understanding. We also provide empirical evidence that exposure to a visual modality does not inherently guarantee improved spatial reasoning, or at least, communication about it, especially in smaller models, which goes against the current assumptions about VLM capabilities and raises questions about how the parametric knowledge gained from the modality is stored and accessed by models. Finally, we show that training lightweight models and small LLMs can match or outperform massive foundation models on these textual grid spatial reasoning tasks, supporting their use as specialized embodied agents.

\section*{Limitations}
This work is, in part, limited by the nature of the proposed tasks. While the simplicity allows for extremely controlled observations on understanding, it is noted that such tasks may not directly generalize to real world settings outside of virtual environments. In particular, embodied scenarios that involve continuous space, visual noise, and partial observability, are not easily represented in the current formulation of GSU. That being said, we think generalizing to continuous space on a grid should be quite feasible, as math is not generally the feature that models struggle the most with. However, having more angles of rotation available to models in \textit{Navigation} tasks will certainly increase the difficulty movement would no longer be nicely axes aligned (though arguable with the egocentric setting the most challenging component is keeping track of headings). In the future, it would be interesting to expand textual problems to incorporate partial observability, exploring if models can tell what would be in their line of sight and reasoning about what may be obscured. It might also present a means of checking information stored during exploration, creating more interesting problems such as reconciling information gathered from different viewpoints (that may not be expressed relative to the same overarching fixed grid). The most direction extension of grid coordinates into more physical problems would likely be in pointclouds and we would be interested to see if our findings generalize to those richer, longer contexts.

Within the confines of the task, we think there is room to further improve smaller models on GSU, explore possibilities such as using performance on a \textit{Follower} task with more direct signal to improve \textit{Instructor} performance, and see if forcing models to communicate about and reason over textual grids can force stronger internal representations or help connect parametric knowledge to visual inforamtion in multimodal models.
%Incorporating pointclouds and other means of parsing the visual modality would also be an important step to generalizing this task.

%This work may also help multimodal models to communicate about and reason over space by forcing stronger internal representations of space in their language components or forcing connections between parametric spatial knowledge and observed multimodal information. Creating models that can reason about space just through text may also support systems that are meant to interface with users in a physical environment that may not be observed (e.g. for privacy concerns).  

\section*{Acknowledgments}
%% LEAVE THIS OUT FOR THE SUBMISSION
%This research used the Delta advanced computing and data resource which is supported by the National Science Foundation (award OAC 2005572) and the State of Illinois. Delta is a joint effort of the University of Illinois Urbana-Champaign and its National Center for Supercomputing Applications.
This work used the Delta system at the National Center for Supercomputing Applications through allocation CIS240333 from the Advanced Cyberinfrastructure Coordination Ecosystem: Services \& Support (ACCESS) program, which is supported by National Science Foundation grants \#2138259, \#2138286, \#2138307, \#2137603, and \#2138296.

All models were used within their licensing agreements described on Huggingface. We thank the Mistral team for access to Mistral-7b and Pixtral, the LLama team for access to Llama-8b and Llama-11b-VL, the Qwen teams for access to  Qwen-7b, Qwen-7b-VL, and Qwen-72b, AI2 for access to Olmo-32b, OpenAI for access to GPT-4o and GPT-OSS-20b, and Google for access to Gemini 3 Pro.

\bibliography{custom}

\appendix
\section{Dataset Generation}
\label{sec:gen_deets}
As GSU is a synthetic dataset and will release generation instructions, some of these statistics may vary or be further adjusted to preference. All dataset statistics reported here are for the test set.

\subsubsection{Navigation}
To generate the Navigation task, we  simulate a grid traversal by randomly generating steps lengths (e.g. how many spaces will be traveled) and randomly choosing a direction from the available options (i.e. in 2d we pick from left/right/forward/back and in 3d we also consider up/down) and pair them together. Then a function determines the final coordinates reached given the list of $(\mathit{direction}, 
\mathit{length})$ pairs and the mode (e.g. \textit{Egocentric} or \textit{Cardinal}). When evaluating the \textit{Follower} role, we provide the $(\mathit{direction}, \mathit{length})$ pairs in natural language and expect a final coordinate as in Figure~\ref{fig:gsu_viz}. In the \textit{Instructor} role, we provide intermediate coordinates along the path as shown in \hyperref[sec:appendix]{Appendix C}. As the base dataset construction is simply $(\mathit{direction}, \mathit{length})$ pairs, the generated datasets are as similar as possible to compare the effects of various settings. That is, the \textit{Instructor} and \textit{Follower} tasks explore the exact same grid traversals but with switched inputs and outputs.

\begin{figure}[h] 
    \centering 
    \includegraphics[width=\linewidth, trim = {0cm 1cm 0cm 0cm}]{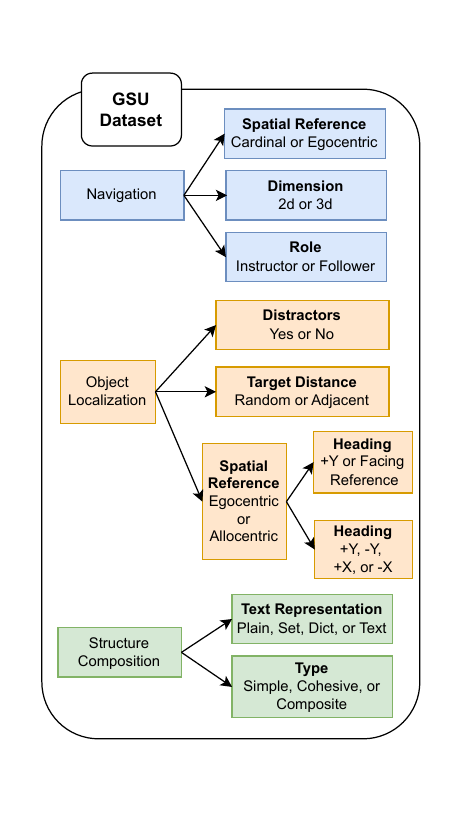}
    \caption{GSU Dataset Tasks and Settings} 
    \label{fig:gsu_dataset}
\end{figure}

Path lengths are 1-4 steps, with 25\% of each length, and an average path length of 2.5. Each step covers 1-10 units, averaging at 5.25. See \ref{tab:direction_stats} for the distribution of movements in each direction. We avoid having multiple steps in the same direction, maximizing the direction changes that make the egocentric task difficult (in the 2D ego task, we change direction in 79.2\% of transitions, in 3D, 49.8\%, in cardinal 91\%).

\begin{table}[h!]
    \centering
    \caption{Navigation Direction Distribution (reported out of 250 possible steps)}
    \label{tab:direction_stats}
    \begin{tabular}{l|c|c|c|c|}
        \hline
        & \multicolumn{2}{c|}{Egocentric} & \multicolumn{2}{c|}{Cardinal} \\
        & 2D & 3D & 2D & 3D \\
        \hline
        Left & 30.4\% & 17.69\% & 21.33\% & 16.67\% \\
        Right & 29.6\% & 18.77\% & 34.67\% & 14.58\% \\
        Front & 20.8\% & 13.72\% & 25.33\% & 12.5\%\\
        Back & 19.2\% & 13.72\% & 18.67\% & 13.54\% \\
        Up & 0 & 20.94\% & 0 & 29.17\% \\
        Down & 0 & 15.16\% & 0 & 13.54\%\\
        
        \hline
    \end{tabular}
\end{table}

\subsubsection{Object Localization}
For the \textit{Egocentric} setting, we randomly sample 2 locations within the grid, which extends $(-20, 20)$ along all axes, designating one the target and one the reference/agent, and a heading along one of the horizontal axes ($+X$, $+Y$, $-X$, $-Y$. For the adjacent case, we also sample one of the coordinates adjacent to the agent location (differing in each dimension by at most 1). In the \textit{Allocentric} setting, we assume that the agent is at the origin and use the two sampled locations as the target and reference blocks. We set the heading either along the $+Y$ axis or towards the reference block. 

For both settings, we select distractor blocks that do not overlap with the target block and programmatically derive the spatial relations.

\textbf{Heading:} In the Object Localization task, we can also vary the direction the embodied agent is facing. For the \textit{Egocentric} setting, facing down a different axis greatly affects the spatial relations (e.g. a block that is in front of an agent facing the $+Y$ direction is to the right of an agent facing $-X$ as in \ref{fig:gsu_viz}). For the \textit{Allocentric} setting, the agent may look in a fixed direction down the $+Y$ axis or directly at the reference object. %Generally, this only makes a difference when the reference and target blocks are equidistant from two axes such that viewing them from either axis leads to different interpretation of their relative placements.

\textbf{Target Distance:}
We choose if the target structures are placed adjacent to the spatial reference (i.e. adjacent to the agent in the \textit{Egocentric} setting and adjacent to the reference block in the \textit{Allocentric} setting) or anywhere on the grid, which is effectively further away due to how few coordinates are adjacent to a single coordinate. Unexpectedly, blocks that are directly adjacent to the reference location are harder for the model to describe spatially (\ref{tab:old_ol}). This trend was observed for the \textit{Allocentric} setting as well, see \ref{tab:adj_vs_rand}. 
% As we limit the grid size to (-20, 20) for each axis, the chance of a randomly placed target being adjacent to the reference is 0.5\% in the 2d setting and 0.04\% in the 3d setting. 

%\textbf{Distractors:} We can include additional blocks of other colors to confuse models. 
We distribute targets across the grid w/ an average location: $(-0.44, 0.99, 0.44)$. In the adjacent \textit{Egocentric OL} setting, the viewer and target are 1.37 apart on average, but may be 13.5 apart in the random setting. In the \textit{Allocentric OL} variant, the target is on average 1.38 from the reference and 13.94 from the viewer (closest 2.4, furthest 22.5 ). For the ground truth answers, there are 1.96 and 1.98 directions per answer for the Ego and Allo variants respectively, with 2 directions being the most common (40\% and 48\% of all answers). However, 1  direction (32\% and 27\%) and 3 directions (28\% and 25\%) are also represented. Table \ref{tab:direction_distribution} shows the distribution of each spatial direction in the ground truth answers.

\begin{table}[h]
\centering
\begin{tabular}{l|c|c|}
\hline
 & \textbf{Egocentric} & \textbf{Allocentric} \\
\hline
Left      & 18.88\% & 18.69\% \\
Right     & 18.37\% & 13.64\% \\
Forward   & 16.84\% & 17.17\% \\
Backward  & 14.80\% & 6.57\%  \\
Up        & 20.41\% & 18.37\% \\
Down      & 14.29\% & 13.64\% \\
\hline
\end{tabular}
\caption{Distribution of Ground Truth Spatial Direction in the Egocentric and Allocentric Object Localization settings}
\label{tab:direction_distribution}
\end{table}

In our prompts (see Appendix \ref{sec:appendix}) we specify that all spatial relations should be given as \textit{target} is \textit{$\langle$spatial relations$\rangle$} of \textit{reference} and we found that nearly all generations followed this prescription. Thus, we found no need to account for the order of objects (i.e. in the \textit{Egocentric} setting we assume all generations are relative to the agent, "the target is to the right of me", and ignore the possibility of "I am to the left of the target").

\subsubsection{Structure Composition}
\label{sec:sc_gen}
To generate the Structure Composition task, we first generate 3 shapes and associated verbal descriptions using a template with minor variations for synonyms and grammatical structure. In the \textit{Simple} setting we select the first shape and associated description, whereas in the \textit{Cohesive} setting we then adjust half the blocks in the structure to be a different color and update the verbal description accordingly. Finally, for the \textit{Composite} setting we translate the 3 shapes to be adjacent and join their textual descriptions with the appropriate spatial terms.

We initially explored a variety of textual representations for coordinates. In our 
\textit{Plain} representation,  coordinates are input into LLMs as sequences resembling pseudo-code: 

\begin{center}
    $\mathit{COLOR}_1 \; X_1 \; Y_1 \; Z_1$ \\
    $...$\\
    $\mathit{COLOR}_N \; X_N \; Y_N \; Z_N$
\end{center}

We also explore alternate representations of coordinates, which this task is particularly suited for since it requires processing far more coordinates than previous tasks. The \textit{Set} representation closely resembles \textit{Plain}, but utilizes structures that the model might have encountered in code: 

"{($\mathit{COLOR}_1$, $X_1$, $Y_1$, $Z_1$), ..., ($\mathit{COLOR}_N$, $X_N$, $Y_N$, $Z_N$)}" 

\textit{Dict} resembles a dictionary where each value is explicitly labeled: 

"{(color = $\mathit{COLOR}_1$, x = $X_1$, y = $Y_1$, z = $Z_1$), ...,  (color = $\mathit{COLOR}_N$, x = $X_N$, y = $Y_N$, z = $Z_N$)}"

\textit{Text} is a more verbal representation and the one used in the previous tasks: 

"a $\mathit{COLOR}_1$ block at ($X_1$, $Y_1$, $Z_1$), ..., and a $\mathit{COLOR}_N$ block at ($X_N$, $Y_N$, $Z_N$)". 

However, for brevity, we chose the overall best performing representation for the main paper, and include the supplementary results here (see \ref{tab:add_cd_res}. We found it interesting that, in general, the \textit{Set} and \textit{Text} representations were the best performing across all models (with notable exceptions, such as Qwen family models highly preferring the \textit{Dict} representation), despite the convention in prior works to use the \textit{Plain} formulation. We also anticipated the \textit{Set} representation performing most similarly to the \textit{Plain} as they were the most similar and were surprised to see that performance was quite different across these two settings. While the \textit{Text} representation is not the most efficient, it follows that textual models would be able to best parse information presented in a form closer to natural language than to a set or dictionary. 

\begin{table*}[h!]
\centering
\begin{tabular}{ll|c|c|c|c|c|c|c|c|c}
\toprule
& & \multicolumn{3}{c|}{Plain}  & \multicolumn{3}{c|}{Dict} &
\multicolumn{3}{c}{Set}\\
% \cmidrule(lr){3-4} \cmidrule(lr){5-6}
& & $ O_s \uparrow$ & $O_c \uparrow$ & $O_f \uparrow$ & $ O_s \uparrow$ & $O_c \uparrow$ & $O_f \uparrow$ & $ O_s \uparrow$ & $O_c \uparrow$ & $O_f \uparrow$  \\
\midrule

\multirow{6}{*}{0-shot}
& Mistral-7b & 15.92 & 42.64 & 12.45 & 13.72 & 47.93 & 10.87 & 15.88 & 46.70 & 13.83 \\
& Pixtral & 37.67 & 74.36 & 14.64 & 44.67 & 70.13 & 13.62 & 32.08 & 72.91 & 7.25\\
& Llama-8b & 21.29 & 63.90 & 13.00 & 22.78 & 63.91 & 15.56 & 25.47 & 68.83 & 18.65\\
& LlamaVL-11b & 35.83 & 69.07 & 11.83 & 24.57 & 60.48 & 12.09 & 20.94 & 62.14 & 13.37\\
& Qwen-7b & 32.17 & 68.72 & 11.48 & 40.03 & 68.32 & 16.73 & 28.52 & 62.76 & 16.65\\
& QwenVL-7b & 18.21 & 60.37 & 8.08 & 33.47 & 52.78 & 12.81 & 30.32 & 45.83 & 11.62 \\
\midrule

\multirow{6}{*}{1-shot}
& Mistral-7b & 27.56 & 58.54 & 13.41 &  32.00 & 55.90 & 13.06 & 38.44 & 71.95 & 11.75\\
& Pixtral & 64.58 & 82.49 & 24.20 & 63.00 & 81.98 & 20.22 & 64.20 & 89.59 & 31.18\\
& Llama-8b & 59.90 & 63.37 & 17.81 & 55.83 & 66.74 & 20.83 & 59.70 & 85.83 & 30.19\\
& Llama-11b & 63.83 & 64.75 & 22.54 & 59.50 & 79.86 & 16.40 & 60.33 & 88.96 & 27.35\\
& Qwen-7b & 50.75 & 76.25 & 16.00 & 63.65 & 81.87 & 23.08 & 54.00 & 88.80 & 29.67\\
& Qwen-7b & 52.52 & 65.72 & 8.24 & 55.67 & 62.57 & 13.38 & 66.90 & 85.27 & 23.37 \\
\midrule

% \multirow{2}{*}{LORA}
% & Mistral-7b & 0.0 & 0.0 & 0.0 & 0.0 & 0.0 & 0.0 \\
% & Llama-8b & 0.0 & 0.0 & 0.0 & 0.0 & 0.0 & 0.0 \\
% % & Qwen-7b & 0.0 & 0.0 & 0.0 & 0.0 & 0.0 & 0.0 & 0.0 & 0.0 & 0.0 & 0.0 & 0.0 & 0.0\\
% \midrule

\multirow{1}{*}{Full FT}
& FlanT5 & 92.92 & 70.32 & 81.59 & 84.25 & 67.27 & 80.88 & 90.08 & 68.07 & 79.68\\
\bottomrule
\end{tabular}
\caption{Structure Composition results for Plain, Dict, and Set Representations}
\label{tab:add_cd_res}
\end{table*}

For the generated block structures, examples have 2-199 blocks, with 61.77 on average. Looking further into the distribution shows a slightly right-skewed distribution, with 35\% of structures having fewer than 15 blocks, and another 35\% having between 15 and 85 blocks, and the remaining 40\% of structures having 85+ blocks. We ensure that within the test set, we represent all 3 structure styles equally, one third being simple, one third being cohesive, and one third being composite. In terms of the other qualities, 49\% of structures are single color, a further 18\% are two-colored, and 33\% have 3 colors. We also have a fairly even distribution of numbers describing the structures, with most between 2-7, but a few outliers with 1 or 8-9. All structures were describable with numbers less than 10 (e.g. no columns or towers were more than 10 blocks high). Most structures could be described in 3 or less numbers , though for a few more might be needed (e.g. a structure with a tower flanked by 2 planes needs at least 6 numbers, more if you want to specify that the plane only has a width of one). Table \ref{tab:sc_dist} further gives the distributions of colors, spatial relations, and shapes. 

\begin{table}[h]
\centering
\caption{Distribution of Spatial Relations, Colors, and Shapes in the Structure Composition task}
\begin{tabular}{ll|c}
\toprule
&  & Percentage \\
\midrule
\multirow{6}{*}{Spatial Relations} & Above  & 19.67\% \\
                 & Below  & 21.2\% \\
                 & Front  & 19.67\% \\
                 & Back   & 10.61\% \\
                 & Left   & 12.12\% \\
                 & Right  & 16.67\% \\
\midrule
\multirow{6}{*}{Colors} & Red    & 16.3\% \\
                 & Orange & 11.96\% \\
                 & Yellow & 21.73\% \\
                 & Green  & 16.85\% \\
                 & Blue   & 17.39\% \\
                 & Purple & 15.76\% \\
\midrule
\multirow{6}{*}{Shapes} & Cube   & 27.7\% \\
                 & Tower  & 24.7\% \\
                 & Row    & 19.88\% \\
                 & Column & 9.64\% \\
                 & Plane  & 18.07\% \\
\bottomrule
\label{tab:sc_dist}
\end{tabular}
\end{table}

\subsection{Shape Metric}
\label{sec:shape_metric}
For the shape metric, we check for synonyms in the generation and map them to their corresponding shape. The synonyms are built around the base vocabulary and then appended to with common outputs in actual generations. For columns, we accept "column", as well any production with "vertical" or "upright" and "line" or "row". For rows, we accept any generation with "row" or "line" that does not mention verticality. For towers (multiple columns grouped together), we accept "tower", "rectangular prism", and "pillar". For planes (rectangular prisms only one block thick), we accept "plane", "platform", "rectangle", "wall", "square", "ring", and "O". For cubes, we accept "cube".

Furthermore, since shapes can be similar to others, if a generation does not match at least one of the ground truth shapes, we offer it partial credit based on how similar the shape it predicted was. Confusing rows and columns leads 0.6 of the full 1.0 received for a total match, as does mixing up a column and a tower. Since towers and cubes are both rectangular prisms, we offer half credit for such mistakes, and 0.1 for towers and planes or planes and cubes, and 0.2 for rows and towers.

\section{Experimental Information}
\subsection{Training Details}
\label{sec:training_deets}

For the LORA fine-tuning runs, we experimented with various ranks, learning rates, and training set sizes. Overall, we found that the model performance plateaued at its highest level with rank 64, a learning rate between 5e-4 and 1e-5 depending on the task, and a training set of size 3000. We used alpha=8 as recommended in the original paper and an AdamW optimizer. We allowed the models to train for 4 epochs, but chose the model with the highest validation accuracy as the one to evaluate. For the FlanT5-large and BART-base runs, we allow the model to train for 15 epochs or until the loss converges using an AdamW optimizer with a learning rate of 5e-5 for Flan and 1e-4 for BART.

This project was run on up to 4 gpuA100x4's using approximately 1500 GPU hours of compute.

\subsection{Ablations}

\begin{table*}[h]
\centering
\resizebox{\textwidth}{!}{
\begin{tabular}{l|c|c|c|c|c|c|c|c|c|c}
\toprule
& \multicolumn{4}{c|}{Follower} & \multicolumn{6}{c}{Instructor} \\
& \multicolumn{2}{c|}{Egocentric 2D} & \multicolumn{2}{c|}{Egocentric 3D} &  \multicolumn{2}{c|}{Cardinal} &  \multicolumn{2}{c|}{Egocentric} &
\multicolumn{2}{c}{Card2Ego} \\
& acc $\uparrow$ & dist. $\downarrow$ & acc $\uparrow$ & dist. $\downarrow$ & acc $\uparrow$ & dist. $\downarrow$ & acc $\uparrow$ & dist. $\downarrow$ & acc $\uparrow$ & dist. $\downarrow$  \\
\midrule

\hline
Flan-T5 w/o finetune & 0 & 34.64 & 0 & 28.28 & 0 & 8.79 & 0 & 7.82 & 0 & 8.23 \\
Flan-T5 w/ finetune & 0.3 & 4.21 & 0.71 & 1.68 & 0.99 & 0.02 & 0.9 & 0.42 & 1.0 & 0 \\
BART w/o finetune & 0 & 34.64 & 0 & 28.28 & 0 & 8.22 & 0 & 7.44 & 0 & 8.21 \\
BART w/ finetune & 0.24 & 2.37 & 0.09 & 7.82 & 0.26 & 2.72 & 0 & 6.29 & 0.85 & 1.37 \\
\hline
\end{tabular}
}
\caption{BART-base and Flan-T5-large Navigation Ablations}
\label{tab:nav_ablations}

\end{table*}

\begin{table*}[h]
\centering
\begin{tabular}{l|c|c||c|c|c|c}
\toprule
& \multicolumn{2}{c||}{Object Localization} & \multicolumn{4}{c}{Structure Composition} \\
& Egocentric & Allocentric & $O_s$ & $O_c$ & $O_f$ & $O_n$  \\
\midrule
\hline
Flan-T5 w/o finetune & 16.5 & 16.67 & 67 & 7.11 & 16.6 & 8.9 \\
Flan-T5 w/ finetune & 19.5 & 94.5 & 99 & 96.17 & 88.83 & 97.07 \\
\hline
\end{tabular}
\caption{BART-base and Flan-T5-Large Ablations over Object Localization and Structure Composition tasks}
\label{tab:ol_sc_ablations}
\end{table*}

Here we show another lightweight model (BART-base) and how both it and Flan-T5 perform before fine-tuning and after for Navigation (Table \ref{tab:nav_ablations}) and Object Localization / Structure Composition (Table \ref{tab:ol_sc_ablations}). We see that performance without any training is quite and improves a great deal for Flan-T5-Large, especially on sequence generation tasks in the \textit{Navigation Instructor} in Table \ref{tab:nav_ablations} and the \textit{Structure Composition} task in Table \ref{tab:ol_sc_ablations}. Since the smaller BART model does not improve as much, we do know that these results may not be generalizable to \textit{all} LMs. Nevertheless, they size difference alone, merits acknowledgement that any LMs can perform comparably to frontier models.

\section{Model Selection}
For the newer models used in the prompting experiments, we considered using the Thinking versions as they might be better suited to reasoning over grid logic, but found that the instruct models performed better. We also explored the Deepseek model distilled on Qwen32b, which received a 0.15 and a 0.26 on the \textit{Egocentric Navigation Follower} task in 2D and 3D (compared to the 0.37 and 0.32 in the standard 32b model shown in \ref{tab:all_res}). It also only received a 0.15 on the \textit{Egocentric Navigation Instructor} task in 3D.

\section{Supplemental Results}
\label{sec:sup}
\subsection{Navigation}
\begin{table*}[h!]
\centering
\small
\begin{tabular}{l|c|c|c|c|c|}
\toprule
& \multicolumn{2}{c|}{Follower} & \multicolumn{3}{c}{Instructor} \\
& Egocentric 2D & Egocentric 3D &  Cardinal & Egocentric & Card2Ego \\
% \cmidrule(lr){3-4} \cmidrule(lr){5-6} \cmidrule(lr){7-8} \cmidrule(lr){9-10}
%\cmidrule{3-8}
\midrule

Mistral-7b & 6.85  & 6.14  & 2.84  & 11.87 & 11.23 \\
\rowcolor{cyan!20} Pixtral & 6.31 & 5.60 & 0.75 & 7.65 & 9.01 \\

Llama-8b & 7.03 & 4.85 & 1.95 & 11.87 & 12.30\\
\rowcolor{cyan!20} LlamaVL-11b & 7.36 & 5.78 & 1.52 & 11.96 & 9.59 \\

Qwen-7b & 22.54 & 20.44 & 1.57 & 8.96 & 7.52 \\
\rowcolor{cyan!20} QwenVL-7b & 7.9 & 12.91 & 4.54 & 7.46 & 7.74\\

GPT-oss-20b & 17.67 & 13.95 & 5.76 & 11.35 & 4.95\\
Olmo-32b & 7.22 & 5.94 & 4.59 & 8.20 & 4.24\\

Qwen-32b & 6.37 & 12.71 & 3.14 & 8.80 & \textbf{4.08} \\
\rowcolor{cyan!20} QwenVL-32b & 6.41 & 5.16 & 6.58 & 14.00 & 4.44 \\

Qwen-72b & \textbf{5.76} & \textbf{3.66} & \textbf{0.0} & \textbf{5.67} & 5.23  \\
\midrule

GPT-4o & 2.94 & 1.40 & 0.56 & 4.04 & 0.77\\
Gemini-3-Pro & \textbf{0.0} & \textbf{0.0} & \textbf{0.0}  & \textbf{0.0}& \textbf{0.0}\\
\midrule

FlanT5 & 4.21 & 1.68 & 0.02 & \textbf{0.42} & \textbf{0.0} \\
Llama-8b & \textbf{1.59} & \textbf{0.17} & 2.58 & 2.31 & 0.21  \\ 
\bottomrule
\end{tabular}

\caption{The distance off scores for each of the \textit{Navigation} experiments in Table \ref{tab:all_res}.}
\label{tab:dists}
\end{table*}

As expected, for the \textit{Navigation Follower} task, we saw much stronger performance in the more predictable \textit{Cardinal} setting (see Table \ref{tab:old_nav}), where even the smaller VLMs (Pixtral, Llama-11b-v, Qwen-7b-v) were consistently solving the 3D task (receiving scores of 0.67, 0.60, and 0.90, respectively), and were extremely proficient on the 2D task (with the LLama and Qwen models getting every example correct). Keeping track of how spatial relations changed with an embodied agents in the \textit{Egocentric} posed a challenge even for GPT-4o and the largest tested Qwen model. 

While the 3D \textit{Cardinal} was more challenging than its 2D counterpart, we suspect that the higher scores in the 3D \textit{Egocentric} setting are due to up/down movements not requiring turns by definition, which effectively shorted the number of rotations and led to fewer opportunities to introduce errors. 

\begin{table}[h!]
\centering
\begin{tabular}{ll|c|c|c|c}
\toprule
&& \multicolumn{2}{c|}{Cardinal 3D} & \multicolumn{2}{c}{Egocentric 3D} \\
&& acc & dist & acc & dist\\
\midrule

\multirow{3}{*}{1}
&Mistral-7b & 0.48 & 4.14 & 0.38 & 6.14 \\
&Llama-8b & 0.62 & 2.65 & 0.47 & 4.85 \\
&Qwen-7b & 0.63 & 7.79 & 0.19 & 20.44 \\
\midrule
\multirow{3}{*}{Few}
&Mistral-7b & 0.19 & 6.23 & 0.17 & 8.32 \\
&Llama-8b & 0.32 & 4.84 & 0.20 & 7.69 \\
&Qwen-7b & 0.50 & 4.31 & 0.29 & 7.07 \\
\bottomrule
\end{tabular}
\caption{1-shot vs Few-shot Navigation: Comparison over 1-shot performance with reasoning vs Few-Shot without reasoning examples}
\label{tab:fsnr}
\end{table}

As explained in Section \ref{sec:exps}, we saw that giving one well-crafted example yielded a greater improvement in performance than several examples with just the ground truth answer. For the \textit{Navigation} task, that is shown in Table \ref{tab:fsnr}, while the improvement against 0-shot prompting can be seen in Table \ref{tab:old_nav}

\begin{table*}[h!]
\centering
\small
\begin{tabular}{ll|c|c|c|c|c|c|c|c|c|c|c|c}
\toprule
& & \multicolumn{8}{c|}{Follower} & \multicolumn{4}{c}{Instructor} \\
%%\cmidrule(lr){3-14}
& & \multicolumn{4}{c|}{Cardinal} & \multicolumn{4}{c|}{Egocentric} & \multicolumn{2}{c|}{Cardinal} & \multicolumn{2}{c}{Egocentric}\\
%%\cmidrule(lr){3-14}
& & \multicolumn{2}{c|}{2d} & \multicolumn{2}{c|}{3d} & \multicolumn{2}{c|}{2d} & \multicolumn{2}{c|}{3d} & 2d & 3d &2d & 3d \\
% \cmidrule(lr){3-4} \cmidrule(lr){5-6} \cmidrule(lr){7-8} \cmidrule(lr){9-10}
%%\cmidrule(lr){3-14}
& & acc.$\uparrow$ & dist$\downarrow$ & acc$\uparrow$ & dist$\downarrow$ & acc$\uparrow$ & dist$\downarrow$ & acc$\uparrow$ & dist$\downarrow$ & acc$\uparrow$ & acc$\uparrow$ & acc$\uparrow$ & acc$\uparrow$ \\
\midrule

\multirow{7}{*}{0-shot}
& Mistral-7b & 0.35 & 6.15 & 0.24 & 14.56 & 0.22 & 9.92 & 0.09 & 17.49 & 0.01 & 0.04 & 0.0 & 0.01 \\
\rowcolor{cyan!20} & Pixtral-12b & 0.85 & 2.39 & 0.54 & 5.34 & 0.29 & 7.34 & 0.34 & 7.05 & 0.21 & 0.10 & 0.09 & 0.07\\
& Llama-8b & 0.81 & 2.34 & 0.35 & 8.52 & 0.33 & 6.86 & 0.36 & 6.55 & 0.05 & 0.11 & 0.02 & 0.02\\
 \rowcolor{cyan!20} & LlamaVL-11b & 0.80 & 2.28 & 0.37 & 9.20 & 0.28 & 7.55 & 0.38 & 6.13 & 0.06 & 0.10 & 0.03 & 0.04\\
& Qwen-7b & \underline{0.96} & \underline{0.49} & \underline{0.86} & \underline{0.85} & \textbf{0.56} & \textbf{4.45} & \textbf{0.53} & \textbf{5.41} & \underline{0.20} & 0.19 & \underline{0.16} & 0.14 \\
\rowcolor{cyan!20} & QwenVL-7b & 0.82 & 2.92 & 0.78 & 3.05 & 0.35 & 9.13 & 0.48 & 6.75 & 0.11 & 0.05 & 0.09 & 0.06\\
& Qwen-72b & - & - & 0.67 & 2.40 &  - & - &  0.36 & 6.21 &  - & \textbf{0.62} &  - & \underline{0.37}\\
\midrule

\multirow{7}{*}{1-shot}
& Mistral-7b & 0.88 & 0.92 & 0.48 & 4.14 & 0.30 & 6.85 & 0.38 & 6.14 & 0.59 & 0.38 & 0.15 & 0.11 \\
 \rowcolor{cyan!20} & Pixtral-12b & 0.97 & 0.37 & 0.67 & 4.20 & \underline{0.33} & \underline{6.31} & 0.50 & 5.60 & \textbf{0.87} & 0.80 & \textbf{0.26} & 0.28 \\
& Llama-8b & 0.99 & 0.04 & 0.62 & 2.65 & \underline{0.33} & 7.03 & 0.47 & 4.85 & 0.69 & 0.54 & 0.06 & 0.23 \\
\rowcolor{cyan!20} & LlamaVL-11b & \textbf{1.00} & \textbf{0.0} & 0.60 & 2.94 & 0.32 & 7.36 & 0.47 & 5.78 & 0.74 & 0.59 & 0.14 & 0.21\\
& Qwen-7b & 0.89 & 3.11 & 0.63 & 7.79 & 0.09 & 22.54 & 0.19 & 20.44 & 0.76 & 0.83 & 0.21 & 0.19\\
\rowcolor{cyan!20} & QwenVL-7b & \textbf{1.0} & \textbf{0.0} & \underline{0.90} & \underline{1.02} & 0.29 & 7.9 & 0.21 & 12.91 & 0.44 & 0.66 & 0.21 & 0.23 \\
& Qwen-72b & - & - & \textbf{1.00} & \textbf{0.00} &  - & - &  \underline{0.75} & \textbf{3.66} &  - & \textbf{0.99} &  - & \underline{0.54} \\
\midrule

% \multirow{2}{*}{LORA}
% & Mistral & 0.0 & 0.0 & 0.0 & 0.0 & 0.0 & 0.0 & 0.0 & 0.0 & 0.0 & 0.0 & 0.0 & 0.0\\
% & Llama & 0.0 & 0.0 & 0.0 & 0.0 & 0.0 & 0.0 & 0.0 & 0.0 & 0.0 & 0.0 & 0.0 & 0.0\\
% & Qwen-7b & 0.0 & 0.0 & 0.0 & 0.0 & 0.0 & 0.0 & 0.0 & 0.0 \\
% \midrule

\multirow{1}{*}{Full FT}
& FlanT5 & - & - & 0.94 & 0.09 & - & - & \textbf{0.82} & 13.14 & - & 0.90 & - & \textbf{0.85} \\
\bottomrule
\end{tabular}
\caption{Navigation task results. We compare 0- and 1-shot prompting on base text-only LLMs, similarly sized VLMs (represented with light blue backgrounds), and  a larger LLM variant of the overall best performing model, as well as a fully fine-tuned LM. We report the accuracies out of 1 for the final coordinates and their distance from the true coordinates for the \textit{Follower} setting and the accuracy out of 1 of the generated instruction for the \textit{Instructor} setting. The best results for each experiment are \textbf{bolded} while the best in each section are \underline{underlined}.}
\label{tab:old_nav}
\end{table*}

\begin{table}[h!]
\centering
\begin{tabular}{|l|c|c|c|c|}
\hline
\textbf{Model Name} & \multicolumn{4}{c|}{\textbf{Sequence Length}} \\ \hline
 & 1 & 2 & 3 & 4 \\ \hline
Qwen-7b & 1.0 & 0.24 & 0.08 & 0.08 \\ \hline
Qwen-7b-v & 0.92 & 0.2 & 0.04 & 0 \\ \hline
Qwen-72b & 1.0 & 0.56 & 0.28 & 0.12 \\ \hline
Chat-GPT & 1.0 & 0.72 & 0.68 & 0.44 \\ \hline
%Row 5 &  &  &  &  \\ \hline
\end{tabular}
\caption{\textit{2D Egocentric Follower Navigation} Performance Deterioration Based on Sequence Length}
\label{tab:models_sequence_length}
\end{table}

At the start of the generated reasoning traces, models seem to understand the standard Cartesian layout (when facing positive y, positive x is to your right), but as they progress through the instruction sequences they fail to keep this consistent with the actions they take. Sometimes, these incorrect justifications are followed by correct steps though, as in: \textit{"we are facing negative x with positive x to the right} [incorrect, $+Y$ would be to the right] \textit{ so that means that turning right means going towards positive x ...} [rest of the step] \textit{... now, we are facing positive y"} [$+Y$ being the correct axes to the right of $-X$, and thus the correct heading upon finishing the step].

For the \textit{Instructor} task, we see that performance is weaker than its \textit{Follower} counterpart, but some models are still able to produce strong instructions sequences that get quite close to the desired endpoint, with Qwen-72b even solving the task completely. But the \textit{Egocentric} setting again breaks what "understanding" these models possess, leading to performances well below a simple model of 875M parameters fully fine-tuned on the problem. Even Qwen-72b and GPT-4o, which show some promise on shorter sequences, get turned around in longer sequences. 

To better understand the source of the errors in the \textit{Egocentric} setting, we proposed the \textit{Card2Ego} setting, which just required understanding how cardinal directions would change relative to an embodied traveller, removing the necessity of reasoning about specific coordinates and where they fall relative to each other. We found that models performed better in this setting, with medium sized models seeing notable improvements, but were still far from completely solving the task. 

Looking at the reasoning traces generated in more detail, we saw that model accuracy dropped on paths with more steps, though this decrease was much more subtle in the cardinal task, occurring more often in weaker or fine-tuned models and due calculation errors / minor mistakes. For egocentric navigation, the dropoff in performance and corresponding increase in the "distance off" metric were much sharper. Many of the weaker models went from accuracies in the 70s and 80s for shorter sequences to 0 or almost 0 in the longer sequences, but even strong models saw performance drops from almost perfect accuracy to numbers in the 60s and 70s. 

Observing the reasoning traces produced by the models for the navigation tasks show that in addition to preferring shorter sequences, most models perform better on backward direction changes as opposed to right/left direction changes. Models were also more likely to produce incorrect conclusion in these longer sequences, particularly in the egocentric navigation tasks. Other times, models across families (GPT, Qwen, Llama) would correctly compute the new headings but not make use of them, falling back to the cardinal version of the problem. Or, they would twist the logic to make the egocentric direction changes align with the direction in a cardinal setting. For example, when facing negative x: \textit{"Moving 8 steps back means moving opposite the direction we are facing. So we will move along positive x, i.e. increase the x value by 8.} [consistent with facing the $-X$ direction] \textit{However, since we are facing negative x, we need to move along negative x, i.e. decrease the x value by 8.} [the complete opposite] \textit{So, our new position is (-3 - 8, 0) = (-11, 0). Since we turned to move, we are now facing negative x and positive y is to our right"} (Llama-8b). This often led models to perform very well on shorter egocentric sequences where the error had not yet compounded, but rapidly diverge on longer sequences. We also found that models produced similar errors within their model family (e.g. Qwen-7b predicting the same incorrect coordinates as the 72B version or the vision version). Smaller models were also less internally consistent and likely to make mistakes about how axes were oriented relative to each other ("since we are facing positive x, negative y is to our left") or other errors (e.g. when moving 9 steps in one direction, incrementing all the x-, y-, z- values by 9). These errors were much less prevalent in the cardinal variants of these tasks, leading us to believe that basic math and reasoning are not the issue in these tasks, but rather keep the rotations / current heading state consistent with previous move and keeping the axes consistent with each other.

\subsection{Object Localization}
For the \textit{Egocentric} Object Localization task, we tested the models on the hardest setting (adjacent target blocks with additional distractors) and in various headings in a 3D grid and saw the expected pattern of improvement from larger models.
%There was, however, no clear correlation between having exposure to a visual modality and improved reasoning over textual coordinate grids in the smaller models. The significant improvement in the Qwen-32b models once vision was added may suggest that VLMs cannot make use of additional spatial information from exposure to visual modalities until the base model reaches a certain size.  %As shown in \todo{reference appendix}, having distractors made the task slightly more challenging for models, as did placing the target blocks adjacent to the references. 

We further looked into accuracy based on the heading (see Table \ref{tab:oo_heading} in the Appendix for more), and found that models mostly performed best in the +$Y$ heading that we would consider the "standard" for the Cartesian grid, with the Qwen family and our fine-tuned model showing the strongest preference (perhaps due to exposure to a similar environment or task in pre-training). For the other 3 headings there was no clear preference between all models, with some preferring the $-Y$ heading as second best and others the headings parallel to the $X$-axis. We see the effects of this axis preference compounded in the \textit{Egocentric} Navigation task, where models struggle to identify spatial directions in non-standard headings.

Especially in the smaller models, we saw a lack of understanding of how axes were situated relative to each other (similar to the Navigation task). 
We also observed confusion over whether the y or z axis was vertical (despite it being explicitly stated) leading to productions such as, \textit{"the block is below and above me"}. There were common incorrect conclusions such as, \textit{"I am facing negative y so smaller values are in front of me.} [correct] \textit{My y is -4 and the block's is 7, so it is \textbf{above} me"}. 
In general, these mistaken deductions were aligned with a standard +$Y$ heading rather than the correct premise preceding them, showing that despite any intermediate generations to the contrary, the models were still operating in that +$Y$ heading.

\begin{table}[h!]
\centering
\begin{tabular}{l|c|c|c|c}
\toprule
& $+Y$ & $+X$ & $-Y$ & $-X$ \\
\midrule
Mistral-7b & 36.90 & 34.98 & 29.13 & 32.26 \\
Llama-8b & 51.33 & 40.23 & 43.80 & 37.84 \\
Qwen-7b & 77.27 & 52.10 & 38.26 & 44.32 \\
\bottomrule
\end{tabular}
\caption{Average Spatial Overlaps $O_s$ in the \textit{Egocentric Object Localization} task broken down by Viewer Heading}
\label{tab:oo_heading}
\end{table}

\begin{table}[h!]
\centering
\begin{tabular}{l|c|c|c|c}
\toprule
& \multicolumn{2}{c|}{$+Y$ Heading} & \multicolumn{2}{c}{Facing Reference} \\
& adj & rand & adj & rand\\
\midrule
0-shot& 32.90 & 38.67 & 29.67 & 38.12 \\
1-shot& 46.50 & 39.83 & 32.32 & 41.75 \\ 
\bottomrule
\end{tabular}
\caption{Adjacent vs Random Block placements in the Allocentric Object Localization Task for Mistral-7b}
\label{tab:adj_vs_rand}
\end{table}

For the \textit{Allocentric} Object Localization task, we evaluated only on the $+Y$ heading so the slight performance drops compared to the \textit{Egocentric} formulation are actually much more pronounced considering that models perform best on this heading. %Only the 3 smallest models performed better 
% Most models performed worse on the \textit{Allocentric} variant, suggesting that models struggled to reason about how two grid coordinates were situated relative to each other from the agent perspective (a third coordinate). Those that performed better were likely benefiting from the lack of distractors in the \textit{Allocentric} setting rather than displaying a particular strength at the task itself.

% We generally saw similar errors on this task, but we did note that with an allocentric spatial reference models confused the forward and backward directions, often flipping them. That is, if the target was between the viewer and the reference they would describe it as behind the reference rather than in front of it as in Figure \ref{fig:qual}.

Looking at the errors made by models for \textit{Egocentric Object Localization}, models that performed better tended to correctly identify how many dimensions differed between the target and viewer coordinates, thus producing the correct number of relative directions. However, poorer performing models tended to predict extra directions (e.g. claiming the target was to the front right of the viewer when it was directly to the right). Looking into these predicted directions, we saw that models were consistently getting up/down difference correct (which makes sense, given that the various headings never affected the z-axis), but tended to predict that objects were to the left when they were to the right, or add in extra above/below directions even when the objects were level with the viewer. The Qwen models had consistently higher performance for the +y heading but other models did not show as much preference. Within the reasoning traces, stronger models were able to break down the problem by each axis and emulate the example reasoning trace to properly using the given heading and surmise what the direction should be. Weaker models were more likely to fail to make a connection between the viewer's heading and the directions, break down the reason trace by axes but proceed as if always facing +y, incorrectly identify the vertical axis, and to make incorrect conclusions from correct premises (e.g. "same y value so the target is level and thus in front of the viewer").

For the \textit{Allocentric} variant, models were more likely to produce fewer dimensions than the ground truth, and very often struggled with the front/back distinction shown in Figure \ref{fig:qual}. Models were mostly likely to extraneously add an incorrect left/right dimensions, and some models seemed to have a preference for which direction they incorrectly predicted (e.g. Qwen72b was more likely to predict left for a ground truth right, than predict right for a ground truth left). We generally saw very structured and consistent reasoning traces, but in the cases of those front/back examples the prompt instructions/examples tended to be ignored in favor of some existing underlying understanding of front/back. Weaker models struggled to keep the axes oriented correctly (often leading to mixups with the vertical dimension) and sometimes referred to where the target was relative to the viewer rather than to the reference.

\begin{table*}[h!]
\centering
\small
\begin{tabular}{ll|c|c|c|c|c|c}
\toprule
& & \multicolumn{4}{c|}{Egocentric} & \multicolumn{2}{c}{Allocentric} \\
& & \multicolumn{2}{c|}{Base} & \multicolumn{2}{c|}{Distractors} & +Y heading & Ref. heading \\
% \cmidrule(lr){3-4} \cmidrule(lr){5-6} \cmidrule(lr){7-8} \cmidrule(lr){9-10}
%\cmidrule{3-8}
& & adj. $\uparrow$ & rand. $\uparrow$ &  adj. $\uparrow$ & rand. $\uparrow$ &  adj. $\uparrow$ &  adj. $\uparrow$  \\
\midrule

\multirow{6}{*}{0-shot}
& Mistral-7b & 30.87 & 34.67 & 30.20 & 31.90 & 32.90 & 29.67  \\
\rowcolor{cyan!20} & Pixtral-12b & 35.22 & 42.54 & 31.42 & 46.73  & 40.73 & 34.43 \\
& Llama-8b & 36.52 & 38.78 & 37.08 & 33.10  & 37.11 & 35.52 \\
 \rowcolor{cyan!20} & LlamaVL-11b & 35.33 & 41.50 & 33.08 & 36.55 & 41.08 & 39.72 \\
& Qwen-7b & \underline{53.55} & \underline{48.20} & \underline{53.07} & \underline{49.02} & \underline{51.75} & \underline{46.82}  \\
\rowcolor{cyan!20} &  QwenVL-7b & 26.18 & 37.83 & 31.98 & 36.15 & 46.75 & 43.25  \\
% & QwenVL-72b & - & - & - & - & - & -  \\
\midrule

\multirow{7}{*}{1-shot}
& Mistral-7b & 30.30 & 40.22 & 30.32 & 39.61 & 46.50  & 32.32 \\
\rowcolor{cyan!20} & Pixtral-12b & \underline{57.55} & \textbf{65.87} & 40.83 & 54.30 & 48.18  & 42.25 \\
& Llama-8b & 48.48 & 52.82 & 46.17 & 50.28 & 50.18 & 42.37  \\
\rowcolor{cyan!20} & LlamaVL-11b & 53.56 & 58.27 & 48.40 & 51.55 & 48.08 & \underline{48.20}  \\
& Qwen-7b & 53.80 & 59.70 & 52.81 & \underline{55.42} & \underline{54.78} & 47.93  \\
\rowcolor{cyan!20} & QwenVL-7b & 51.78 & 52.42 & 48.67 & 52.42 & 44.95 & 41.08  \\
& QwenVL-72b & - & - & \underline{79.53} & \textbf{79.45} & 53.83 & 47.62 \\
\midrule

\multirow{2}{*}{LORA}
& Mistral-7b & - & - & 36.72 & 62.62 & 59.03 & 47.89 \\
& Llama-8b & - & - & \textbf{80.05} & \underline{72.60} & \underline{61.58} & \underline{84.03}  \\
% & Qwen-7b & 0.0 & 0.0 & 0.0 & 0.0 & 0.0 & 0.0 & 0.0 & 0.0 \\
\midrule

\multirow{1}{*}{Full FT}
& FlanT5 & \textbf{98.08} & 65.75 & 73.03 & 62.50 & \textbf{100} & \textbf{96.1} \\
\bottomrule
\end{tabular}
\caption{Object Localization spatial overlap, $O_s$, scores across the \textit{Egocentric} and \textit{Allocentric} settings, as well various headings and distractor settings. VLM counterparts are in light blue backgrounds, best results overall are \textbf{bolded} and best results in generation setting are \underline{underlined}.}
\label{tab:old_ol}
\end{table*}

\begin{table*}[h!]
\centering
\small
\begin{tabular}{ll|c|c|c|c|c|c}
\toprule
& & \multicolumn{3}{c|}{Set}  & \multicolumn{3}{c}{Text}  \\
% \cmidrule(lr){3-4} \cmidrule(lr){5-6}
& & $ O_s \uparrow$ & $O_c \uparrow$ & $O_f \uparrow$ & $ O_s \uparrow$ & $O_c \uparrow$ & $O_f \uparrow$  \\
\midrule

\multirow{3}{*}{1-shot}
&Mistral-7b & 26.14 & 54.12 & 15.18 & 34.73 & 57.10 & 24.22 \\
& Llama-8b & 59.98 & 63.58 & 18.69 & 60.26 & 67.05 & 22.16 \\
& Qwen-7b & 54.70 & 74.86 & 20.85 & 57.18 & 71.78 & 19.14 \\
\midrule

\multirow{3}{*}{Few-shot}
&Mistral-7b& 38.44 & 71.95 & 11.75 & 49.75 & 72.60 & 26.33 \\
&Llama-8b & 59.70 & 85.83 & 30.19 & 62.04 & 86.69 & 26.10 \\
& Qwen-7b & 54.00 & 88.80 & 29.69 & 54.58 & 85.43 & 25.67 \\

\bottomrule
\end{tabular}
\caption{1-shot vs Few-shot Structure Composition: Comparison over 1-shot performance with reasoning vs Few-Shot without}
\label{tab:fsnr2}
\end{table*}

\subsection{Structure Composition}
We saw spatial results similar to the \textit{Allocentric} ones, with just slight improvements in some models, likely due to the perspective difference in the task definition: in the \textit{Allocentric} setup we treat the reference as having its own spatial reference (i.e. it has a left and right of its own, mirroring those of the viewer), whereas in the Structure Composition setup we defer to a viewer facing along the $+Y$ axis for spatial terms to preserve the simplicity of the task. This simpler spatial referencing is offset by the higher number of blocks and the need to group and then orient them correctly. 

Naturally, the $O_f$ subtask was the most difficult since it requires models to match groupings of coordinates to real-life shapes that exist outside of grid and coordinate settings, making the fully fine-tuned model's performance all the more impressive. Reflective of the wide vocabluary that models might apply to these shapes, we used generous synonyms in the shape metric (see Appendix \ref{sec:shape_metric}). As such, an $O_f$ score around 50, reflects a mastery of simple shapes like columns and rows in their individual and \textit{Composite} settings and some limited success with more complex shapes such as towers, rectangular prisms, and cubes, whereas one near 90 suggests a mastery of individual shapes and composite structures, with occasional errors, usually on larger shapes. Finally, we did see shape preference in the model outputs, with certain models generally not predicting certain types of shapes (depending on the model, this was generally a cube or a tower). 

We also show results for the various coordinate representation formats discussed in Section \ref{sec:sc_gen} in Table \ref{tab:add_cd_res} and the performance difference between more few-shot examples with no reasoning and a single 1-shot example with strong reasoning in Table \ref{tab:fsnr2}.

\onecolumn
\section{Prompt and Generation Examples}
\label{sec:appendix}

\subsection{Instructor Navigation}
\begin{tcolorbox}[width=\columnwidth,colback={white},title={2d Egocentric Navigation Instructor Prompt w/ 1-shot Example},colbacktitle=pink,coltitle=black]
You are in a 2D environment with (x, y) coordinates set up like a standard horizontal Cartesian plane. You will start at the origin (0, 0), which is at the center of this grid. You are currently facing the positive y direction, with the positive x direction to your right. However, when you move in a direction you must turn to face that direction, rotating your frame of reference. For example, if you move left, you will rotate 90 degrees and be facing the negative x direction with positive y to your right.

First, let me give you an example!
You start at (0, 0), go to (3, 0), go to (3, 1), and end at (3, -1). Describe the path that traverses the provided coordinates.

I must remember that each time I move left, right, or back, I will be turning to face a new direction.

1. To get from (0, 0) to (3, 0), I must move 3 steps to the right. I will now be facing the positive x direction, with negative y to my right.

2. To get from (3, 0) to (3, 1), I must move 1 step left. I will now be facing positive y, with positive x to my right.

3. To get from (3, 1) to (3, -1), I must move 2 steps back.
So, my path is [ANS] right 3, left 1, back 2 [/ANS].
Great, now let's try a real problem!

Start at (0, 0, 0). Go to (0, 7, 0). Go to (0, -1, 0). End at (-4, -1, 0). Describe the path that you will take to traverse the provided coordinates. Format your answer as a series of directions and distances, e.g. [ANS] forward 2, right 3, back 1 [/ANS]. Let's go!

\end{tcolorbox} 

\begin{tcolorbox}[width=\columnwidth,colback={white},title={Mistral-7b Generation },colbacktitle=yellow,coltitle=black]

\textbf{CORRECT ANSWER :}  ['forward 7', ' backward 8', ' right 4']

\textbf{GENERATED ANSWER :}  ['forward 7', 'backward 8', 'backward 4', 'forward 1']

\end{tcolorbox}

\begin{tcolorbox}[width=\columnwidth,colback={white},title={2d Card2Ego Navigation Instructor Prompt w/ 1-shot Example},colbacktitle=pink,coltitle=black]
You are on a 2D grid and will be given a path using cardinal directions (North, East, South, West) that you need to convert into egocentric directions (left, right, forward, backward). Keep in mind that to move in a cardinal direction, you must turn to face it. This means that the egocentric instructions will not map directly to cardinal ones, but change depending on the direction you last moved. E.g. if you just moved East and then want to move South, then the egocentric instruction is to move right, since South is to your right if you're facing East.

Let's begin with an example. You start by facing North and the path is: West 2, North 3, East 1.

The first step is easy. I am facing North and must move West which is to my left. So the first step is: left 2.

Now, for the second step I am facing the direction I just moved in, West, and I want to move North. If I am facing West, North is to my right. So the instruction is right 3.

Then, for the third step I am facing North and want to go East 1. East is to my right and so I need to move right 1.

Finally, I can put that all together for: [ANS] left 2, right 3, right 1 [/ANS]
Now, you give it a try! You start facing North and the path is: West 3, East 8, South 1, South 10. What is the path expressed with egocentric directions?

\end{tcolorbox} 

\begin{tcolorbox}[width=\columnwidth,colback={white},title={GPT-4o Generation },colbacktitle=yellow,coltitle=black]

\textbf{CORRECT ANSWER :}  [left 3, backward 8, right 1, forward 10]

\textbf{GENERATED ANSWER :}  
Let's break it down step by step.

1. Start facing North. To go West means turning left.
   - Instruction: left 3

2. Now facing West. To go East means turning around (backward).
   - Instruction: backward 8

3. Now facing East. To go South means turning right.
   - Instruction: right 1

4. Still facing South. To go South again means going forward.
   - Instruction: forward 10

Putting it all together: [ANS] left 3, backward 8, right 1, forward 10 [/ANS]

\end{tcolorbox}

\subsection{Follower Navigation}
\begin{tcolorbox}[width=\columnwidth,colback={white},title={3d Egocentric Navigation Follower Prompt w/ 1-shot Example},colbacktitle=pink,coltitle=black]
You are in a 3D environment with (x, y, z) coordinates set up like a standard Cartesian plane. The x and y dimensions are horizontal, while the z dimension is the vertical component. You will start at the origin (0, 0, 0), which is at the center of this grid. You are currently facing the positive y direction, with the positive x direction to your right. However, when you move in a direction you must turn to face that direction, rotating your frame of reference. For example, if you move left, you will rotate 90 degrees and be facing the negative x direction with positive y to your right.
Explain your final coordinates after travelling. Please format your final coordinates as: [ANS] (x, y, z) [/ANS]. Let's go!

Let's start with an example: First, you move 3 steps to your right. Next, you move 2 steps down. You move 4 steps backwards. Finally, you move 2 steps left. Where are you now?

To solve this, we should break down the steps we take.

1. We start at (0, 0, 0) facing the positive y direction, with positive x to our right.

2. Moving 3 steps right means moving along positive x, i.e. increasing the x value by 3. So, our new position is (3, 0, 0). Also, we turned to face positive x, so now negative y is to our right.

3. Moving 2 steps down means moving along negative z, i.e. decreasing the z value by 2. So, our new position is (3, 0, -2). Remember that moving up and down doesn't change our heading, so we are still facing positive x with negative y to our right.

4. Moving 4 steps back means moving opposite the direction we are facing. So we will move along negative x, i.e. decrease the x value by 4. So, our new position is (-1, 0, -2). Since we turned to move, we are now facing negative x and positive y is to our right.

5. Moving 2 steps left means moving away from the right. So we will move along negative y, i.e. decrease y by 2. So, our new position is (-1, -2, -2).
Our final position is [ANS] (-1, -2, -2) [/ANS]

Now, let's try a real problem!
First, you move 7 steps to your right. You then move 5 steps forward. You move 10 steps up. Lastly, you move 5 steps backward. Explain your final coordinates: 

\end{tcolorbox} 

\begin{tcolorbox}[width=\columnwidth,colback={white},title={Mistral-7b Generation},colbacktitle=yellow,coltitle=black]
\textbf{Correct Answer is :}  {'x': 7, 'y': 0, 'z': 10}

\textbf{Generated Coordinates are :}  (7, 0, 10)

\end{tcolorbox} 

\subsection{Egocentric Object Localization}
\begin{tcolorbox}[width=\columnwidth,colback={white},title={3d Egocentric Object Localization w/ Distractors, 1-shot Example},colbacktitle=pink,coltitle=black]
You are in a 3D environment with (x, y, z) coordinates set up like a standard Cartesian plane. The x and y dimensions are horizontal, while the z dimension is the vertical component. All axes range from (-10, 10). Your task is to describe where objects are relative to you, without using coordinates. Instead use relative descriptions like 'directly behind me' or 'to my left'. Explain your thought process and please format your final answer with [ANS] tags like so: the green block is [ANS] in front of me and to my left [/ANS]. Let's get started!
 
Here is an example: You are at (2, 2, 0), facing the negative y direction, so negative x is to your right. The positive z axis is always up. There is a blue block at (-3, -1, 0). Where is the blue block relative to you? 

For the x dimension, I know that negative x is to my right. That means x values smaller than mine are to my right and ones bigger than mine are to my left. My x coordinate is 2 and the blue block's is -3. So it is to my right.

Next, for the y dimension, I know that I am facing negative y, so it is in front of me. This means that y values smaller than mine are in front of me and those bigger than mine are behind me. My y coordinate is 2 and the blue is at -1. So, it is in front of me.

Now, for the z dimension, my orientation does not matter. Higher z values are above me and lower z values are below me. However, we are both at z = 0, so we are level. Putting that together, the blue block is [ANS] to my right and slightly to the front [/ANS]

Now let's try a real problem! You are at (-1, 5, -9), facing the negative y direction, so negative x is to your right. The positive z axis is always up. There is a yellow block at (-2, -1, -3), a orange block at (3, 2, -3), a green block at (-1, 4, 10), a red block at (-7, 1, 7), and a blue block at (2, 9, -9). Where is the red block relative to you?

\end{tcolorbox} 

\begin{tcolorbox}[width=\columnwidth,colback={white},title={Generation by Qwen-7b },colbacktitle=yellow,coltitle=black]

\textbf{Correct Answer is :} The red block is [ANS] in front of me, to my right, and above me [/ANS]

\textbf{Generated Description is :} The red block is [ANS] to my left and behind me [/ANS]

\textbf{BLEU :}  0.0

\textbf{BERT :} 0.8915

\textbf{SPATIAL OVERLAP :}  0.0

\end{tcolorbox} 

\subsection{Allocentric Object Localization}
\begin{tcolorbox}[width=\columnwidth,colback={white},title={3d Allocentric Object Localization w/ +Y heading, 1-shot Example},colbacktitle=pink,coltitle=black]

You are in a 3D environment with (x, y, z) coordinates set up like a standard Cartesian plane. The x and y dimensions are horizontal, while the z dimension is the vertical component.All axes range from (-20, 20). Your task is to describe where objects are relative to you and other objects, without using coordinates. Instead use relative descriptions like 'directly in front of the blue cylinder'. Explain your thought process and please format your final answer with [ANS] tags like so: the yellow block is [ANS] below and to the back right of [/ANS] the purple block. Let's get started!

You are at the origin. There is a red block at (0, 8, -7) and a blue block at (0, 7, -8). You may assume that you are facing the positive y direction. The positive z axis is always up. Where is the blue block relative to the red block given your point of view?

\end{tcolorbox} 

\begin{tcolorbox}[width=\columnwidth,colback={white},title={Llama-8b Generation },colbacktitle=yellow,coltitle=black]

\textbf{Correct Answer is :} The red block is [ANS] below and in front of [/ANS] the blue block.

\textbf{Generated Description is :} The red block is [ANS] below and to the back of [/ANS] the blue block.

\textbf{BLEU :}  0.0

\textbf{BERT :}  0.826

\textbf{SPATIAL OVERLAP :}  0.33

\end{tcolorbox}

\subsection{Composite Description}
\begin{tcolorbox}[width=\columnwidth,colback={white},title={Composite Structure Composition Task},colbacktitle=pink,coltitle=black]
You are in a 3D grid environment with (x, y, z) coordinates set up like a standard Cartesian plane. The x and y dimensions are horizontal, while the z dimension is the vertical component. Your task is to describe a set of blocks to someone without mentioning coordinates or axes, instead describe the structures as a whole. Format your answer with [ANS] tags like so: there are [ANS] 6 orange blocks in a column [/ANS]. 

% Let's start with an example. 
% Blocks:
% The blocks placed on the grid are:
% {(color : blue, x : 1, y : 1, z : 0), (color : blue, x : 1, y : 1, z : 1), (color : blue, x : 1, y : 1, z : 2), (color : blue, x : 1, y : 1, z : 3)}
% These blocks form [ANS] a blue column 4 blocks high [/ANS].
% Now, try a real problem.

Blocks:
The blocks placed on the grid are:
{(color : purple, x : 0, y : 0, z : 0), (color : purple, x : 0, y : 0, z : 1), (color : purple, x : 0, y : 0, z : 2), (color : purple, x : 0, y : 0, z : 3), (color : purple, x : 0, y : 0, z : 4), (color : purple, x : 0, y : 0, z : 5), (color : purple, x : 0, y : 0, z : 6), (color : purple, x : 0, y : 0, z : 7), (color : purple, x : 0, y : 1, z : 0), (color : purple, x : 0, y : 1, z : 1), (color : purple, x : 0, y : 1, z : 2), (color : purple, x : 0, y : 1, z : 3), (color : purple, x : 0, y : 1, z : 4), (color : purple, x : 0, y : 1, z : 5), (color : purple, x : 0, y : 1, z : 6), (color : purple, x : 0, y : 1, z : 7), (color : purple, x : 1, y : 0, z : 0), (color : purple, x : 1, y : 0, z : 1), (color : purple, x : 1, y : 0, z : 2), (color : purple, x : 1, y : 0, z : 3), (color : purple, x : 1, y : 0, z : 4), (color : purple, x : 1, y : 0, z : 5), (color : purple, x : 1, y : 0, z : 6), (color : purple, x : 1, y : 0, z : 7), (color : purple, x : 1, y : 1, z : 0), (color : purple, x : 1, y : 1, z : 1), (color : purple, x : 1, y : 1, z : 2), (color : purple, x : 1, y : 1, z : 3), (color : purple, x : 1, y : 1, z : 4), (color : purple, x : 1, y : 1, z : 5), (color : purple, x : 1, y : 1, z : 6), (color : purple, x : 1, y : 1, z : 7), (color : yellow, x : 0, y : -2, z : 0), (color : yellow, x : 0, y : -2, z : 1), (color : yellow, x : 0, y : -2, z : 2), (color : yellow, x : 0, y : -2, z : 3), (color : yellow, x : 0, y : -2, z : 4), (color : yellow, x : 0, y : -1, z : 0), (color : yellow, x : 0, y : -1, z : 1), (color : yellow, x : 0, y : -1, z : 2), (color : yellow, x : 0, y : -1, z : 3), (color : yellow, x : 0, y : -1, z : 4), (color : yellow, x : 1, y : -2, z : 0), (color : yellow, x : 1, y : -2, z : 1), (color : yellow, x : 1, y : -2, z : 2), (color : yellow, x : 1, y : -2, z : 3), (color : yellow, x : 1, y : -2, z : 4), (color : yellow, x : 1, y : -1, z : 0), (color : yellow, x : 1, y : -1, z : 1), (color : yellow, x : 1, y : -1, z : 2), (color : yellow, x : 1, y : -1, z : 3), (color : yellow, x : 1, y : -1, z : 4), (color : red, x : 0, y : 0, z : 8), (color : red, x : 0, y : 1, z : 8), (color : red, x : 0, y : 2, z : 8), (color : red, x : 0, y : 3, z : 8), (color : red, x : 0, y : 4, z : 8), (color : red, x : 1, y : 0, z : 8), (color : red, x : 1, y : 4, z : 8), (color : red, x : 2, y : 0, z : 8), (color : red, x : 2, y : 4, z : 8), (color : red, x : 3, y : 0, z : 8), (color : red, x : 3, y : 1, z : 8), (color : red, x : 3, y : 2, z : 8), (color : red, x : 3, y : 3, z : 8), (color : red, x : 3, y : 4, z : 8)}
Now describe the structure they made. 

\end{tcolorbox}

\begin{tcolorbox}[width=\columnwidth,colback={white},title={Qwen-7b Generation },colbacktitle=yellow,coltitle=black]

\textbf{Correct Answer is :}  8 x 2 x 2 purple tower with a tower of yellow blocks 5 high and 2 wide in front of it and a empty 4 x 5 red wall above it

\textbf{Generated Description is :}  two large purple platforms stacked on top of each other, both extending from the bottom to the top of the visible area, and a smaller yellow platform that extends from slightly below the bottom to just above the middle of the visible area. Additionally, there is a red block formation at the very top, creating a cross-like pattern centered around the central vertical axis

\textbf{BLEU :}  0.0

\textbf{BERT :}  0.8357

\textbf{SPATIAL OVERLAP :}  0.2

\textbf{COLOR OVERLAP :}  1.0

\textbf{SHAPE OVERLAP :}  0.4

\textbf{NUMERICAL OVERLAP :} 0.0

\end{tcolorbox}

\subsection{Combination Task}
\label{sec:combo_task}
\begin{tcolorbox}[width=\columnwidth,colback={white},title={Combination Navigation and Structure Localization Task },colbacktitle=pink,coltitle=black] 

You are in a 3D environment with (x, y, z) coordinates set up like a standard Cartesian plane. The x and y dimensions are horizontal, while the z dimension is the vertical component.You are at the origin (0, 0, 0), which is at the center of this grid, facing the positive y direction, with the positive x direction to your right. However, when you move in a direction you must turn to face that direction, rotating your frame of reference. For example, if you move left, you will rotate 90 degrees counterclockwise and be facing the negative x direction with positive y to your right.
There is a red block at (-10, 9, -5), \textbf{[excluded for human readability]} and a red block at (-12, 11, -4).
You move 5 steps to your right. You then move 10 steps backward. You move 5 steps to your left. You also move 2 steps to your left. Next, you move 3 steps backward. You move 3 steps forward as well. Next, you move 1 step backward. Finally, you move 6 steps to your left.
Now, where is the row relative to the plane given your point of view?

\end{tcolorbox}

\begin{tcolorbox}[width=\columnwidth,colback={white},title={Gemini Generation },colbacktitle=yellow,coltitle=black]

To determine the position of the row relative to the plane from your point of view, we first need to trace your movements to find your final position and facing direction.

**1. Tracing Your Movements:**

* **Initial State:** You start at `(0, 0, 0)` facing the positive y-direction (+y). Positive x (+x) is to your right. 

* **Move 5 steps to your right:** You turn to face +x and move 5 steps. 

  * *Position:* `(5, 0, 0)` | *Facing:* +x

* **Move 10 steps backward:** "Backward" from +x is -x. You turn to face -x and move 10 steps.

  * *Position:* `(-5, 0, 0)` | *Facing:* -x
  
* **Move 5 steps to your left:** Left of -x is -y. You turn to face -y and move 5 steps.

  * *Position:* `(-5, -5, 0)` | *Facing:* -y
  
* **Move 2 steps to your left:** Left of -y is +x. You turn to face +x and move 2 steps.

  * *Position:* `(-3, -5, 0)` | *Facing:* +x

* **Move 3 steps backward:** Backward from +x is -x. You turn to face -x and move 3 steps.

  * *Position:* `(-6, -5, 0)` | *Facing:* -x

* **Move 3 steps forward as well:** "Forward" means you continue moving in your currently facing direction (-x).

  * *Position:* `(-9, -5, 0)` | *Facing:* -x

* **Move 1 step backward:** Backward from -x is +x. You turn to face +x and move 1 step.

  * *Position:* `(-8, -5, 0)` | *Facing:* +x

* **Move 6 steps to your left:** Left of +x is +y. You turn to face +y and move 6 steps.

  * *Position:* `(-8, 1, 0)` | *Facing:* +y

**Your Final State:** You are at **(-8, 1, 0)** facing the positive y-direction (**+y**).
From your perspective:

* **Forward:** +y direction

* **Right:** +x direction

* **Up:** +z direction

**2. Locating the Structures:**

By analyzing the coordinates of the blocks, we can identify the specific grouped structures:

* **The Plane (Frame):** There is a hollow rectangular plane of blocks situated exactly at `z = 8`. It spans `x = 0` to `5` and `y = 7` to `13`.

  * *Center of the Plane:* `(2.5, 10, 8)`
  
* **The Row:** There is a horizontal line of 7 blocks that spans entirely along the x-axis from `x = -3` to `3`.

  * *Center of the Row:* `(0, 8, -5)`

**3. Determining Relative Position from Your Perspective:**

Now, we compare the center of the Row `(0, 8, -5)` to the center of the Plane `(2.5, 10, 8)` relative to your point of view facing +y.

* **Left/Right (x-axis):** The row's center `x = 0` is less than the plane's center `x = 2.5`. Since +x is your right, a smaller x coordinate means it is to the **left**.

* **Front/Behind (y-axis):** The row's center `y = 8` is closer to your position at `y = 1` than the plane's center at `y = 10`. Because you are facing +y, objects with a smaller y-value are closer to you, placing the row **in front of** the plane. 

* **Above/Below (z-axis):** The row is at `z = -5` while the plane is up at `z = 8`. Since your up is +z, the row is vertically **below** the plane.

**Conclusion:**

Given your point of view, the row is located to the **left, in front of, and below** the plane.

\textbf{SPATIAL OVERLAP :}  1.0

\end{tcolorbox}

\end{document}